\documentclass{article}

\usepackage[preprint]{neurips_2025}

\usepackage{dsfont}
\usepackage[utf8]{inputenc} 
\usepackage[T1]{fontenc}    
\usepackage{hyperref}       
\usepackage{url}            
\usepackage{booktabs}       
\usepackage{amsfonts}       
\usepackage{nicefrac}       
\usepackage{microtype}      
\usepackage{xcolor}         
\usepackage{graphicx}       
\usepackage{amsmath}        
\usepackage{amsthm}        
\usepackage{cleveref}
\usepackage{multirow}
\usepackage{subcaption}
\usepackage{xcolor}
\usepackage{colortbl}
\usepackage{capt-of,etoolbox}
\usepackage{wrapfig}
\usepackage[linesnumbered,ruled,vlined]{algorithm2e}

\definecolor{tabhighlight}{HTML}{e5e5e5}
\crefname{algocf}{Algorithm}{Algorithms}

\newtheorem{theorem}{Theorem}[section]

\title{Multi-Modal Dataset Distillation in the Wild}

\author{
  Zhuohang Dang$^{1}$\quad  Minnan Luo$^{1}$\quad Chengyou Jia$^{1}$ \\
  \textbf{Hangwei Qian}$^{2}$\quad \textbf{Xiaojun Chang}$^{3}$\quad \textbf{Ivor W. Tsang}$^{2}$\\
  $^1$Xi'an Jiaotong University, $^2$A*STAR, $^3$ University of Science and Technology of China\\
  dangzhuohang@stu.xjtu.edu.cn
  \vspace{-5mm}
}
\usepackage{xspace}

\makeatletter
\DeclareRobustCommand\onedot{\futurelet\@let@token\@onedot}
\def\@onedot{\ifx\@let@token.\else.\null\fi\xspace}

\def\eg{\emph{e.g}\onedot} 
\def\ie{\emph{i.e}\onedot} 
 
\def\etc{\emph{etc}\onedot} \def\vs{\emph{vs}\onedot}

\makeatother

\begin{document}

\maketitle

\begin{abstract}
    Recent multi-modal models have shown remarkable versatility in real-world applications.
    However, their rapid development encounters two critical data challenges.
    First, the training process requires large-scale datasets, leading to substantial storage and computational costs.
    Second, these data are typically web-crawled with inevitable noise, \ie, partially mismatched pairs, severely degrading model performance.
    To these ends, we propose \textbf{M}ulti-modal dataset \textbf{D}istillation in the \textbf{W}ild, \ie, \textbf{MDW}, the first framework to distill \textbf{\textit{noisy}} multi-modal datasets into compact \textbf{\textit{clean}} ones for effective and efficient model training.
    Specifically, MDW introduces learnable fine-grained correspondences during distillation and adaptively optimizes distilled data to emphasize correspondence-discriminative regions, thereby enhancing distilled data's information density and efficacy.
    Moreover, to capture robust prior knowledge on cross-modal correspondence from noisy real data, MDW proposes dual-track collaborative learning to address the risky data noise, alleviating information loss with certifiable noise tolerance.
    Extensive experiments validate MDW's theoretical and empirical efficacy with remarkable scalability, surpassing prior methods by over 15\% across various compression ratios, highlighting its appealing practicality for applications with diverse efficacy and resource needs.
\end{abstract}

\section{Introduction}
Recent advancements in multi-modal models, like Gemini \cite{team2023gemini} and GPT-4o \cite{hurst2024gpt}, have shown remarkable versatility in real-world applications.
However, such models' training process is confronted with significant data challenges.
First, these models rely on large-scale training datasets, leading to substantial data storage and model training costs \cite{yu2023dataset}. 
Moreover, these data are automatically collected by sourcing co-occurring sample pairs from the Internet, which inevitably contain imperfect alignment due to ubiquitous noise in real-world scenarios, \ie, partially mismatched pairs (PMPs).
For example, CC3M \cite{sharma2018conceptual}, comprising 3 million real-world image-text pairs, contains around 30\% PMPs that severely mislead model training \cite{huang2021learning}.
These challenges impede the participation of practitioners with limited resources, thereby restricting the rapid development and broad application of this field.

\begin{wrapfigure}{r}{0.5\textwidth}
  \centering
  \vspace{-3mm}
  \includegraphics[width=0.48\textwidth]{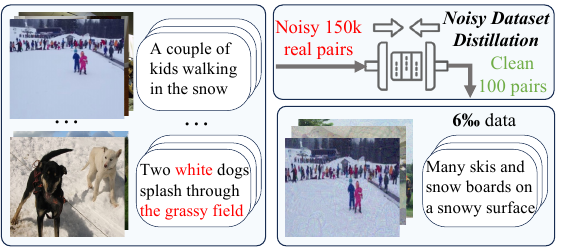}
\caption{Illustration of noisy multi-modal dataset distillation with 1500:1 compression ratio.}
\vspace{-3mm}
\label{fig:intro}
\end{wrapfigure}

Recent efforts have made progress in partially addressing these challenges. Pioneering works \cite{lors,vl_distill} explore dataset distillation in the multi-modal domain, aiming to create compact training datasets while preserving model performance. Although these methods alleviate dependence on large-scale data, they implicitly assume ideally well-matched datasets and struggle with PMPs, leading to distillation collapse. 
Therefore, as shown in \Cref{fig:intro}, we pose the challenging problem: \textbf{\emph{can we distill the large-scale noisy multi-modal dataset into a compact clean one for effective and efficient model training?}}

Intuitively, multi-modal dataset distillation can be viewed as generating and enhancing multi-modal correspondence patterns within distilled data to facilitate model training.
To deal with PMPs, the intuitive solution is a two-step pipeline: first filtering PMPs, then distilling the remaining clean data. 
However, such filtration-based methods excessively exclude samples to ensure the purity of clean data, leading to severe information loss from the original dataset.
On the other hand, existing methods \cite{vl_distill, lors} treat all pixels in distilled images equally during distillation, overlooking the visual information redundancy in RGB space for multi-modal learning \cite{liang2023factorized}. 
In practice, only a small portion of image pixels is related to cross-modal correspondence, while the rest is largely redundant or even distracting.
Therefore, the equality among pixel-level updates hinders the targeted enhancement of correspondence-discriminative regions, severely diminishing the distillation efficacy.

To address these issues, we explore the more challenging real-world scenario and propose a novel framework, \textbf{MDW}, for noisy multi-modal dataset distillation.
MDW proposes \textbf{fine-grained correspondence-enhanced distillation} to improve distilled data's information density and efficacy, as well as \textbf{dual-track collaborative learning} (DTCL) to address risky PMPs when capturing multi-modal priors from noisy real data.
In detail, unlike similarity mining \cite{lors} with isolated sample pairs, MDW introduces soft matching probability to capture holistic relationships across all samples, yielding better alignment with the image-text contrastive learning paradigm.
Moreover, to enhance multi-modal correspondence patterns' refinement, MDW identifies correspondence-discriminative regions and emphasizes them for more updates during distillation, thereby reducing redundant areas' impact.
Additionally, to mine robust correspondence patterns from real data, DTCL complementarily explores negative matches across all samples, beyond conventional positive learning after noisy sample filtration.
Especially, negative matches' abundance makes their non-correspondence relationships inherently noise robust, offering reliable supervision to enhance dataset distillation and alleviate real data's information loss.
We theoretically validate DTCL's efficacy by proving its noise tolerance.

We conduct extensive experiments on various benchmarks, \eg, Flickr30K and COCO with synthetic noise (0\%, 30\% and 50\%) and CC3M with real-world noise.
Compared to existing methods, datasets distilled by our MDW improve the performance of models trained from scratch by 15\% on standard test sets, validating MDW's empirical and theoretical efficacy in noisy multi-modal dataset distillation.
Additionally, our MDW demonstrates superior scalability and adaptability across architectures, underscoring its promising potential and applicability in real-world scenarios.

Our contributions are summarized as follows:
\textbf{(1)} We propose MDW, the first attempt at multi-modal dataset distillation in noisy real-world scenarios, for reducing data requirements while ensuring robustness against prevalent PMPs in real-world data.
\textbf{(2)} MDW enhances the distilled dataset's information density and efficacy with fine-grained correspondence-enhanced distillation strategy. Moreover, MDW introduces dual-track collaborative learning to capture multi-modal priors from noisy real data with certifiable noise tolerance.
\textbf{(3)} Extensive experiments validate MDW's theoretical and empirical efficacy, surpassing prior methods across various settings by more than 15\%.

\section{Related Work}
\noindent\textbf{Dataset Distillation.}
Current methods typically fall into two types:  
(1) Meta-learning based methods \cite{nguyendataset,zhou2022dataset,loo2023dataset} typically employ a bi-level optimization strategy, with the inner loop training on the distilled data and the outer loop optimizing the meta datasets.
(2) Matching-based methods synthesize distilled datasets by matching training by-products (\eg, gradients \cite{zhao2021dataset}, features \cite{wang2022cafe} or trajectories \cite{cazenavette2022dataset}) between surrogate models trained on the distilled and real datasets.
Although effective, such methods primarily focus on the image domain, limiting their application in multi-modal learning. 

Recently, VL-Distill \cite{vl_distill} pioneers the multi-modal dataset distillation by jointly matching visual and textual encoders' trajectories. LoRS \cite{lors} further employs a similarity mining strategy to synthesize the similarity among distilled data. However, these methods struggle with partially mismatched pairs (PMPs) prevalent in real-world data, leading to distillation collapse.  
In contrast, our MDW effectively alleviates PMPs' negative impact, greatly boosting its applicability in real-world scenarios.

\noindent\textbf{Multi-Modal Learning.}
Multi-modal learning has shown remarkable versatility across real-world applications \cite{jia2024generating,jia2024agentstore,jia2024chatgen,ping2025autogps,wu2024atlas}.
Current methods \cite{lee2018stacked, diao2021similarity} encode inputs into a unified feature space to align matching pairs while distinguishing mismatched ones.
However, these methods implicitly assume perfect data alignment and are vulnerable to PMPs prevalent in real-world.
To address this, \cite{huang2021learning,han2023noisy,yang2023bicro,yang2024robust} explore neural networks' memorization effect to develop noise-robust strategies to identify and suppress PMPs during training.
However, these methods fail to explore the fine-grained correspondence within samples and suffer from severe information loss from the real data due to trivial noise sample filtration.
Conversely, our MDW explores fine-grained correspondence-enhanced distillation and mines reliable supervision even within noisy subsets for enhanced distillation efficacy.

\section{Methodology}
\subsection{Problem Formulation}
We follow \cite{lors,vl_distill} to adopt the image-text retrieval task as a proxy to investigate the noisy multi-modal dataset distillation. Let $\mathcal{D}=(\mathcal{V},\mathcal{T},\mathcal{Y})$ denote an image-text dataset, where $\mathcal{V}=\{V_i\}_{i=1}^{N}$ and $\mathcal{T}=\{T_i\}_{i=1}^{N}$ are image and text sets with $N$ training samples.
 $\mathcal{Y}=\{y_{ij}\}_{i,j=1}^{N}$ is the matching label set, with $y_{ij}=1$ \emph{iff} $V_i$ and $T_j$ are matched. Ideally, sample pairs with the same indices are matched, \ie, $y_{ii}=1, \forall i \in [1,N]$.
However, due to the pervasive noise during data collection, an unknown proportion $\eta$ of matching labels $\mathcal{Y}$ are corrupted, resulting in a noisy multi-modal dataset $\mathcal{D}_\eta=(\mathcal{V},\mathcal{T},\hat{\mathcal{Y}})$.
Here, we assume the noise is uniform and independent, \ie, \( y_{ii} \) is corrupted from 1 to 0 with probability \( \eta \), while \( y_{ij}, \forall j \neq i \) is corrupted from 0 to 1 with probability \( \frac{\eta}{N-1} \).

Given noisy dataset $\mathcal{D}_\eta$ and $M\ll N$, we seek to learn a compact clean synthetic dataset $\tilde{\mathcal{D}}=(\tilde{\mathcal{V}},\tilde{\mathcal{T}},\tilde{\mathcal{P}})$ with $M$ samples.
Here, $\tilde{\mathcal{P}}\in\mathbb{R}^{M\times M}$ denotes the soft matching probability matrix to capture the fine-grained, sample-level correspondences among distilled data, to enrich the information density of $\tilde{\mathcal{D}}$.
Moreover, $\tilde{\mathcal{V}}$, $\tilde{\mathcal{T}}$ and $\tilde{\mathcal{P}}$ are regarded as learnable parameters and optimized based on the knowledge from noisy dataset $\mathcal{D}_\eta$. 
Our goal is to enable models trained from scratch on \(\tilde{\mathcal{D}}\) to achieve generalization performance comparable to those trained on the clean dataset \(\mathcal{D}\).

\begin{wrapfigure}{r}{0.575\textwidth}
    \vspace{-10mm}
    \centering
    \includegraphics[width=0.565\textwidth]{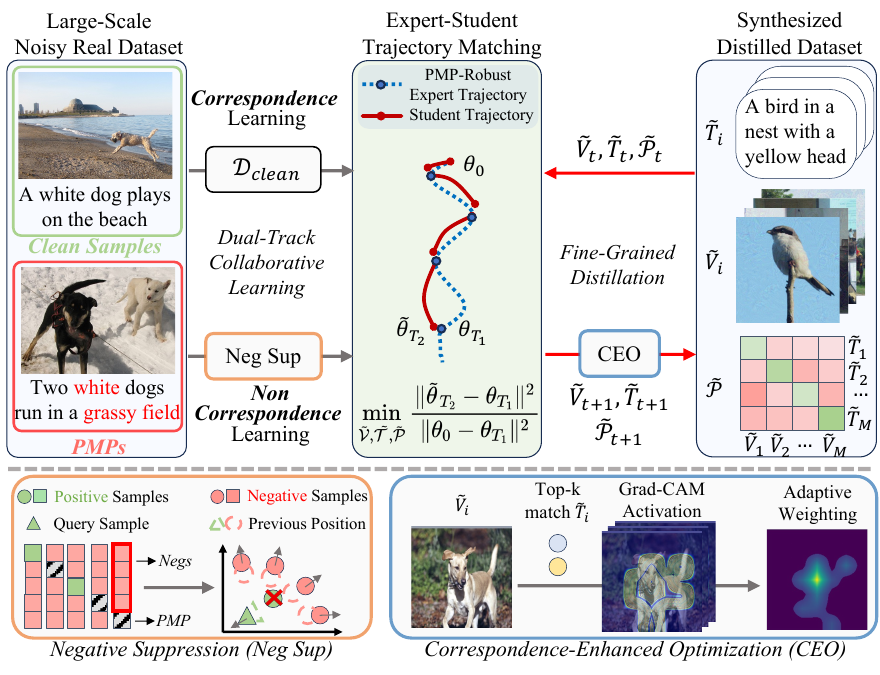}
    \vspace{-1.5mm}
    \caption{Overview of our MDW. The upper part illustrates the expert-student trajectory matching with fine-grained correspondence-enhanced distillation, while the lower part depicts non-correspondence learning and CEO strategy.}
    \label{fig:overview}
    \vspace{-2.5mm}
  \end{wrapfigure}

\subsection{Model Overview}
\Cref{fig:overview} presents MDW, the first framework for noisy multi-modal dataset distillation, built on a long-term model trajectory matching framework akin to \cite{lors,vl_distill}.
Specifically, MDW simultaneously trains an \textit{expert} model on the real dataset and a \textit{student} model on the distilled dataset with identical training settings to facilitate knowledge transfer.
The core idea is to achieve dataset distillation by implicitly matching the training parameter trajectory \cite{cazenavette2022dataset} of expert and student models.
In this sense, MDW proposes fine-grained correspondence-enhanced distillation, refining learnable matching probabilities and adaptively optimizing distilled data to emphasize correspondence-discriminative regions, thereby enhancing distilled data's efficacy.
To capture multi-modal priors from noisy real data with certifiable noise tolerance, MDW introduces dual-track collaborative learning (DTCL) that complementarily utilizes non-correspondence relationships within negative matches across all samples, beyond conventional correspondence learning with positive matches after sample filtration.
The following sections detail each component.

\subsection{Fine-Grained Correspondence-Enhanced Distillation}
To begin with, we initialize the distilled dataset $\tilde{\mathcal{D}}=\{\tilde{\mathcal{V}},\tilde{\mathcal{T}},\tilde{\mathcal{P}}\}$ by randomly selecting $M$ samples from the real dataset $\mathcal{D}_\eta$, \ie, $\tilde{\mathcal{V}}=\{\tilde{V}_i\}_{i=1}^M$, $\tilde{\mathcal{T}}=\{\tilde{T}_i\}_{i=1}^M$ and $M\ll N$.
Moreover, we introduce a soft matching probability matrix, $\tilde{\mathcal{P}}\in\mathbb{R}^{M\times M}$ to replace hard matching labels, and initialize it as the identity matrix.
Then we detail how to optimize $\tilde{V}_i$, $\tilde{T}_i$ and $\tilde{\mathcal{P}}$ with the knowledge of $\mathcal{D}_\eta$ as follows.

Given a batch of multi-modal inputs $\{V_i, T_j\}_{i,j=1}^B$, we employ modality-specific encoders $f$ and $g$ to project the visual and textual inputs into a unified feature space, \ie, $F_V^i=f(V_i)$ and $F_T^j=g(T_j)$.
The similarity of image-text pair $(V_i,T_j)$ is computed using a similarity function $h$ as $h(F_V^i, F_T^j)$. In this paper, $h$ is cosine similarity function and we denote $h(F_V^i, F_T^j)$ as $h_{ij}$ or $h(V_i,T_j)$ for brevity.

\noindent\textbf{Fine-Grained Distillation.}
We employ the long-term expert-student trajectory matching framework \cite{vl_distill,cazenavette2022dataset} for effective knowledge transfer from real dataset $\mathcal{D}_\eta$ to distilled dataset $\tilde{\mathcal{D}}$.
Assume the expert and student models are parameterized by \( \theta \) and \( \tilde{\theta} \), respectively. We initialize both models with any same parameters, \ie, \( \theta_{0} = \tilde{\theta}_{0} \).
Moreover, we train the expert and student models on the $\mathcal{D}_\eta$ and $\tilde{\mathcal{D}}$ with identical training settings and symmetric image-to-text ($i2t$) and text-to-image ($t2i$) losses:
{\small
  \begin{align}
    L_c=-\sum\nolimits_{i=1}^B(y_{ii}^{i2t}\log& p_{ii}^{i2t}+y_{ii}^{t2i}\log p_{ii}^{t2i}),
    \text{with}\ p_{ij}^{i2t}=\frac{e^{h_{ij}/\tau}}{\sum_{k=1}^Be^{h_{ik}/\tau}}, p_{ij}^{t2i}=\frac{e^{h_{ij}/\tau}}{\sum_{k=1}^Be^{h_{kj}/\tau}} \label{eq:l_c},
  \end{align}}
where $\tau$ is the temperature parameter; $y_{ii}$ are the matching labels.
Here, \( y_{ii} \) is hard labels in real datasets, \ie, \( y_{ii}^{i2t} = 1 \). Conversely, it is defined as soft labels for student training on distilled datasets, \ie, $y_{ii}^{i2t} = \frac{e^{\tilde{p}_{ii}/\tau}}{\sum_{k=1}^{B}e^{\tilde{p}_{ik}/\tau}}$. 
Compared to the conventional \textit{\textbf{one-to-one hard}} modality correspondence, the soft matching probability $\tilde{\mathcal{P}}$ effectively captures the \textit{\textbf{complex, fine-grained, sample-level}} correspondence among distilled data, thereby greatly enriching the information density of $\tilde{\mathcal{D}}$.

Finally, after performing \( T_1 \) and \( T_2 \) gradient descent updates on the expert and student models, we obtain the updated parameter sets \( \theta_{T_1} \) and \( \tilde{\theta}_{T_2} \).
The $t$-th iteration of distilled dataset $\tilde{\mathcal{D}}_t=\{\tilde{\mathcal{V}}_t,\tilde{\mathcal{T}}_t,\tilde{\mathcal{P}}_t\}$ is updated by matching the accumulated difference between student and expert trajectory, \ie,
\begin{equation}
\label{eq: update}
  L=\frac{\|\tilde{\theta}_{T_2} - \theta_{T_1} \|^2}{\|\theta_{0}-\theta_{T_1} \|^2}, \tilde{\mathcal{V}}_{t+1}=\tilde{\mathcal{V}}_t-\alpha_{v}\frac{\partial L}{\partial \tilde{\mathcal{V}}_t}, \tilde{\mathcal{T}}_{t+1}=\tilde{\mathcal{T}}_t-\alpha_{t}\frac{\partial L}{\partial \tilde{\mathcal{T}}_t}, \tilde{\mathcal{P}}_{t+1}=\tilde{\mathcal{P}}_t-\alpha_{p}\frac{\partial L}{\partial \tilde{\mathcal{P}}_t},
\end{equation}
where $\alpha_v$, $\alpha_t$ and $\alpha_p$ are the learning rates for $\tilde{\mathcal{V}}$, $\tilde{\mathcal{T}}$ and $\tilde{\mathcal{P}}$, respectively.

\noindent\textbf{Correspondence-Enhanced Optimization.}
Intuitively, \Cref{eq: update} treats all pixels equally during $\tilde{\mathcal{V}}$'s optimization. However, \Cref{fig:overview} shows that only a small subset of pixels contributes to cross-modal correspondence due to substantial visual redundancy in the RGB space \cite{liang2023factorized}. 
Therefore, we propose to adaptively modulate the update magnitude of each pixel, thereby promoting more effective generation and refinement of cross-modal correspondence patterns in the distilled image.

We begin by identifying correspondence-relevant pixels in the distilled image. 
Inspired by Grad-CAM's localization of class-discriminative regions for visual classification \cite{selvaraju2017grad}, we reformulate image-text matching as a batch-wise classification task. 
Specifically, for a given distilled image $\tilde{V}_i$, we regard its paired distilled text $\tilde{T}_i$ as the label and the similarity score $h(\tilde{V}_i, \tilde{T}_i)$ as the classification logit.
Consequently, we extract the feature maps $F^k$ from the last convolutional layer of expert's visual encoder and compute the Grad-CAM activation map $M^{ii}$ for $(\tilde{V}_i,\tilde{T}_i)$, \ie,
\begin{equation}
\alpha_{ii}^{k} = \frac{1}{H \times W} \sum\nolimits_{h} \sum\nolimits_{w} \frac{\partial h(\tilde{V}_i, \tilde{T}_i)}{\partial F^k_{h, w}}, \quad M^{ii} = \operatorname{ReLU}(\sum\nolimits_{k=1}^C\alpha_{ii}^{k} F^k),
\end{equation}
where $\alpha_{ii}^{k}$ is a weight that measures the importance of $F^k$ for the target distilled text $\tilde{T}_i$.
Moreover, since our MDW introduces a soft matching probability $\tilde{\mathcal{P}}$, $\tilde{V}_i$ can naturally capture multiple cross-modal correspondence patterns associated with different distilled texts $\tilde{T}_j$. Therefore, we select the top-$k$ texts with the highest matching probabilities from $\tilde{\mathcal{P}}$ for the given $\tilde{V}_i$ and compute the corresponding Grad-CAM activation maps $M^{ij}$. The final activation map $M^i$ is then obtained by averaging these $k$ maps, effectively aggregating correspondence cues across texts to comprehensively model the relevance strength between pixels and cross-modal correspondence patterns.

Given $M^{i}$, we consider pixels with activation values above the mean $\mu(M^{i})$ as correspondence-discriminative, whose optimization will be prioritized to better refine cross-modal correspondence patterns. To this end, we introduce a weighting strategy that adaptively assigns greater importance to these pixels, \ie, the weighting factor $A^{i}_{j,k}$ for each pixel $(j,k)$ in $M^{i}$ is formulated as:
\begin{equation}
A^{i}_{j,k} = \mathds{1}[M^{i}_{jk}>\mu(M^{i})][e^{\beta(M^{i}_{jk}-\mu(M^{i}))}] + 1; \tilde{{V}}_{i,t+1}=\tilde{V}_{i,t}-\alpha_{v}A^{i}\odot \frac{\partial L}{\partial \tilde{V}_{i,t}},
\end{equation}
where $\mathds{1}$ is the indicator function evaluating whether the pixel $(j,k)$ is correspondence-discriminative and $\odot$ denotes the element-wise product. The hyperparameter $\beta$ modulates the sharpness of the weighting function, effectively emphasizing pixels with higher activation values.
Moreover, we apply an exponential moving average (EMA) update to $A^{i}$ to aggregate historical weighting information, thereby promoting consistent and stable correspondence patterns enhancement during optimization.

Ideally, after training multiple randomly initialized expert and student models and matching their trajectories, our fine-grained correspondence-enhanced distillation can effectively encode knowledge from real data to small-scale distilled data. 
However, such iterations of the expert-student trajectory matching process are vulnerable to PMPs in real data, which mislead experts to capture erroneous multi-modal priors and result in distillation collapse.
Therefore, it is essential to address the impact of risky PMPs on expert model training during each iteration of expert-student trajectory matching.

\subsection{Mining Robust Multi-Modal Priors Against PMPs in the Wild}
To address the risky PMPs in real data $\mathcal{D}_\eta$, it is intuitive to divide $\mathcal{D}_\eta$ into clean and noisy subsets \cite{huang2021learning}, then distilling only the clean portion.
However, such a solution struggles with the information insufficiency, as the strict criteria for ensuring clean subset purity excessively exclude potentially clean samples \cite{xia2023regularly}.
This leads to severe training instability and substantial information loss in $\mathcal{D}_\eta$, diminishing distilled data's efficacy.
To this end, we propose dual-track collaborative learning to complementarily mine reliable information from noisy samples for enhanced expert training.

\noindent\textbf{Dual-Track Collaborative Learning.}
To begin with, we follow \cite{huang2021learning} to identify noisy samples in $\mathcal{D}_\eta$ based on neural networks' memorization effects, wherein model prioritizes learning clean samples before fitting noisy ones.
Therefore, we can divide $\mathcal{D}_\eta$ into clean and noisy subsets, where noisy samples can be identified with small-loss criterion (Please refer to Appendix \ref{sec: PMP_filtration} for more details).

Subsequently, we employ \Cref{eq:l_c} to encode the modality correspondence knowledge from clean subsets without the negative impact of PMPs.
Note that it is risky to directly utilize positive matches in noisy subsets during expert training, as it contains a high proportion of PMPs. 
In contrast, due to the abundance of negative matches for a given sample, there is a low likelihood of matching label corruption for negative matches, \ie, $\eta\ \vs\ \frac{\eta}{N-1}$.
Consequently, the non-correspondence relationship within negative matches, \ie, \textit{the image and text are irrelevant}, provides a robust supervision signal against noise.
Based on this observation, we propose a complementary approach that guides the model in leveraging such modality non-correspondence relationships, \ie,
\begin{equation}
  L_n=-\sum\nolimits_{i}\sum\nolimits_{j\neq i}(\log(1-p_{ij}^{i2t}) + \log(1-p_{ij}^{t2i})).
  \label{eq: l_n}
\end{equation}
During student training on the distilled data,  $L_n$ is combined with soft matching probability to capture the fine-grained modality non-correspondence relationships among negative matches. The final loss function is formulated as \( L = L_c + L_n \) to jointly guide the training of both expert and student models.

Theoretically, we prove that $L_n$ is robust to corrupted matching labels to highlight its efficacy, \ie,
\begin{theorem}
  Given the noise ratio $\eta<\frac{N-1}{N}$, $L_n$ is noise tolerant against uniform corrupted matching label $\hat{y}$ in instance-level image-text matching task.
\label{th: noise robustness}
\end{theorem}
In essence, $L_n$ ensures model convergence to a solution that deviates from the optimal one obtained with clean matching labels by a bounded margin.
Unlike traditional noise-tolerant MAE loss $L_{mae}$ \cite{ghosh2017robust}, $L_n$ is more effective for small-scale datasets, particularly in the context of distilled data. 
While $L_{mae}$ treats samples equally for noise tolerance, it inadvertently downplays informative hard samples that are crucial for effective learning, thereby increasing training difficulty on small-scale distilled datasets \cite{lin2017focal,zhang2018generalized}.
Conversely, $L_n$ adaptively emphasizes informative hard samples with larger gradients, striking a better balance between noise tolerance and learning efficacy.
Moreover, $L_n$'s noise tolerance facilitates negative matches' reliable utilization to enhance expert, enabling more precise noisy sample filtration \cite{huang2021learning} and thus collaboratively enhancing the correspondence learning.

\section{Experiments}

\subsection{Experimental Settings}
\noindent\textbf{Datasets.} We adopt three widely used benchmarks for thorough evaluations of our MDW. \textbf{Flickr30K} \cite{young2014image} and \textbf{COCO} \cite{lin2014microsoft} are image-text matching datasets with 31k and 123k images, each paired with five textual descriptions. Following \cite{vl_distill}, we adopt 29k/1k/1k and 113k/5k/5k splits for training/validation/testing.
\textbf{CC3M} \cite{sharma2018conceptual} contains 3M images with a single caption each. As crawled from the Internet, CC3M naturally contains around 30\% PMPs \cite{huang2021learning}. For our experiments, we use a subset of CC3M with 104k samples, namely CC104K. Specifically, we randomly select 100k images from the training split for training, 1k/3k images from the validation split for validation/testing.

\noindent\textbf{Evaluation Metrics.} After noisy multi-modal dataset distillation, we train multiple models with the distilled data from scratch and evaluate them on the corresponding test set. We report the average efficacy of the standard retrieval metric Recall@K (R@K), measuring the proportion of correct matches in the top-K retrieved samples. We systematically report R@1/5/10 in both image-to-text and text-to-image retrieval scenarios, which are aggregated as R\_sum to evaluate overall efficacy.

\begin{table*}[!t]
	\newcommand{\tabincell}[2]{\begin{tabular}{@{}#1@{}}#2\end{tabular}}
	\centering
	\caption{
		Distilling noisy Flickr30K and COCO into 100 samples. Our MDW and PMP-robust LoRS variants (introduced as strong baselines in this work) are highlighted in gray.
	Model on original dataset training achieves R@1/5/10 of 33.9/65.1/75.2 (Image) and 27.3/57.1/69.7 (Text) on Flickr30K; while 19.6/45.6/59.5 (Image) and 16.9/41.9/55.9 (Text) on COCO.
	}
	\vspace{-3mm}

	\label{table:flicker}
	\resizebox{\textwidth}{!}{ 
		\begin{tabular}{c|c|ccc|ccc|c|ccc|ccc|c}
		\toprule[1.5pt]
		\multirow{3}{*}{Noise}&\multirow{3}{*}{Methods}&\multicolumn{7}{c|}{Flickr30K}&\multicolumn{7}{c}{MS-COCO 1K}\\
		&&\multicolumn{3}{c|}{Image$\longrightarrow$Text}&\multicolumn{3}{c|}{Text$\longrightarrow$Image}&&\multicolumn{3}{c|}{Image$\longrightarrow$Text}&\multicolumn{3}{c|}{Text$\longrightarrow$Image}&\\
		\cline{3-16}
			& &R@1&R@5&R@10&R@1&R@5&R@10&R\_Sum&R@1&R@5&R@10&R@1&R@5&R@10&R\_Sum\\
			\midrule
            \multirow{5}{*}{0\%}&K-center& 0.6   & 5.0  & 7.6 & 0.7 & 3.1  & 6.1 & 23.1  & 1.4 & 3.7  & 5.5 & 0.4 & 1.4 & 2.5 & 14.9\\
			&Forget& 1.2   & 4.2  & 9.7 & 0.7 & 2.4  & 5.6 & 23.8  & 0.7 & 2.6  & 4.8 & 0.3 & 1.5 & 2.5 & 12.4\\
			&vl-distill& 9.9   & 28.3  & 39.1 & 4.7 & 15.7  & 24.6 & 122.3  & 2.5 & 10.0  & 15.7 & 1.3 & 5.4 & 9.5 & 44.4\\
			&LoRS& 11.8   & 35.8  & 49.2 & 8.3 & 24.1  & 35.1 & 164.3  & 3.3 & 12.2  & 19.6 & 1.8 & 7.1 & 12.2 & 56.2\\
			&Ours& \textbf{15.7}   & \textbf{40.8} & \textbf{55.3} & \textbf{9.5} & \textbf{28.6}  & \textbf{40.2} & \textbf{190.1}  & \textbf{5.0} & \textbf{17.7}  & \textbf{27.7} & \textbf{2.8} & \textbf{10.8} & \textbf{17.7} & \textbf{81.7}\\
			\midrule
			\multirow{7}{*}{30\%}&K-center& 1.1 & 3.8  & 6.7 & 0.6 & 2.7  & 4.4 & 19.3 &0.5 & 1.9  & 3.8 & 0.2 & 1.0 & 1.9 & 9.3\\
			&Forget&  0.8  &  3.9 & 5.9  & 0.5 & 2.5  & 4.3 & 17.9  &0.4 & 1.8  & 3.3 & 0.2 & 0.9 & 1.7 & 8.3\\
			&vl-distill& 7.2 & 23.7 & 37.1 & 3.5 &  11.2 & 17.7 & 100.4  & 2.1 & 8.4  & 14.1 & 1.0 & 4.4 & 7.9 & 37.9\\
			&LoRS& 10.9   & 31.5  & 43.5 & 7.0 & 20.6  & 31.9 & 145.4  & 2.6 & 10.1  & 17.5 & 1.3 & 5.5 & 9.9 & 46.9 \\
			&\cellcolor{tabhighlight}LoRS-C& \cellcolor{tabhighlight}11.7   &\cellcolor{tabhighlight}33.7  &\cellcolor{tabhighlight}48.3 &\cellcolor{tabhighlight}7.5 &\cellcolor{tabhighlight}22.7  &\cellcolor{tabhighlight}33.6 &\cellcolor{tabhighlight}157.5  &\cellcolor{tabhighlight}2.8 &\cellcolor{tabhighlight}11.5  &\cellcolor{tabhighlight}18.7 &\cellcolor{tabhighlight}1.6 &\cellcolor{tabhighlight}6.0 &\cellcolor{tabhighlight}10.5 &\cellcolor{tabhighlight}51.1 \\
			&\cellcolor{tabhighlight}LoRS+& \cellcolor{tabhighlight}10.7   &\cellcolor{tabhighlight}31.6  &\cellcolor{tabhighlight}47.6 &\cellcolor{tabhighlight}7.3 &\cellcolor{tabhighlight}21.9  &\cellcolor{tabhighlight}32.7 &\cellcolor{tabhighlight}151.8  &\cellcolor{tabhighlight}2.7 &\cellcolor{tabhighlight}10.6  &\cellcolor{tabhighlight}18.0 &\cellcolor{tabhighlight}1.4 &\cellcolor{tabhighlight}5.9 &\cellcolor{tabhighlight}10.3 &\cellcolor{tabhighlight}48.9 \\
			&\cellcolor{tabhighlight}Ours& \cellcolor{tabhighlight}\textbf{14.3}   & \cellcolor{tabhighlight}\textbf{39.1}  &\cellcolor{tabhighlight}\textbf{53.7} &\cellcolor{tabhighlight}\textbf{8.6} &\cellcolor{tabhighlight}\textbf{26.2}  &\cellcolor{tabhighlight}\textbf{38.7} &\cellcolor{tabhighlight}\textbf{180.6}  &\cellcolor{tabhighlight}\textbf{4.9} &\cellcolor{tabhighlight}\textbf{16.4}  &\cellcolor{tabhighlight}\textbf{25.8} &\cellcolor{tabhighlight}\textbf{2.6} &\cellcolor{tabhighlight}\textbf{9.6} &\cellcolor{tabhighlight}\textbf{16.0} &\cellcolor{tabhighlight}\textbf{75.3}\\
			\midrule
			\multirow{7}{*}{50\%}&K-center& 0.8 &  2.4 & 4.9 & 0.4 & 1.7  & 3.0 & 13.2  & 0.2 & 1.5  & 2.8 & 0.1 & 0.9 & 1.6 & 7.1 \\
			&Forget& 0.8  & 2.8  & 5.1 & 0.3 & 1.8 & 3.6 & 14.4  & 0.3  & 1.3  & 2.6 & 0.1 & 0.8 & 1.5 & 6.6 \\
			&vl-distill& 6.3   & 18.1  & 26.4  & 3.2 & 11.0  & 18.6 & 83.6  & 1.7 & 7.0 & 12.4 & 0.9 & 4.2 & 7.2 & 33.4 \\
			&LoRS& 10.4   & 27.3  & 39.9 & 5.9 & 19.3  & 29.4 & 132.2  & 2.4 & 8.8  & 14.6 & 1.1 & 4.3 & 7.9 & 39.1\\
			&\cellcolor{tabhighlight}LoRS-C& \cellcolor{tabhighlight}11.2   &\cellcolor{tabhighlight}32.8  &\cellcolor{tabhighlight}46.8 &\cellcolor{tabhighlight}7.3 &\cellcolor{tabhighlight}21.1  &\cellcolor{tabhighlight}32.9 &\cellcolor{tabhighlight}152.1  &\cellcolor{tabhighlight}2.6 &\cellcolor{tabhighlight}10.9  &\cellcolor{tabhighlight}18.5 &\cellcolor{tabhighlight}1.3 &\cellcolor{tabhighlight}5.4 &\cellcolor{tabhighlight}9.3 &\cellcolor{tabhighlight}48.0\\
			&\cellcolor{tabhighlight}LoRS+& \cellcolor{tabhighlight}10.6  &\cellcolor{tabhighlight}30.9  &\cellcolor{tabhighlight}44.8 &\cellcolor{tabhighlight}6.2 &\cellcolor{tabhighlight}19.4  &\cellcolor{tabhighlight}31.2 &\cellcolor{tabhighlight}143.1  &\cellcolor{tabhighlight}2.5 &\cellcolor{tabhighlight}9.9  &\cellcolor{tabhighlight}16.5 &\cellcolor{tabhighlight}1.2 &\cellcolor{tabhighlight}5.1 &\cellcolor{tabhighlight}8.9 &\cellcolor{tabhighlight}44.1\\
            &\cellcolor{tabhighlight}Ours& \cellcolor{tabhighlight}\textbf{13.8}   & \cellcolor{tabhighlight}\textbf{38.4}  &\cellcolor{tabhighlight}\textbf{50.4} &\cellcolor{tabhighlight}\textbf{8.4} &\cellcolor{tabhighlight}\textbf{25.9}  &\cellcolor{tabhighlight}\textbf{37.3} &\cellcolor{tabhighlight}\textbf{174.2}  &\cellcolor{tabhighlight}\textbf{4.6} &\cellcolor{tabhighlight}\textbf{14.8}  &\cellcolor{tabhighlight}\textbf{23.7} &\cellcolor{tabhighlight}\textbf{2.3} &\cellcolor{tabhighlight}\textbf{9.2} &\cellcolor{tabhighlight}\textbf{15.5} &\cellcolor{tabhighlight}\textbf{70.1} \\
			\bottomrule[1.5pt]
	\end{tabular}}
 \vspace{-7mm}
\end{table*}

\noindent\textbf{Implementation Details.}
Following \cite{vl_distill}, we adopt Normalizer-Free ResNet (NFNet) \cite{brock2021high} pretrained on ImageNet \cite{deng2009imagenet} as the image encoder and a pretrained BERT-base \cite{devlin2018bert} with an appended linear layer as the text encoder. 
For efficiency, the text encoder is frozen during training and the expert model is trained on the full real dataset for 10 epochs 20 times as pre-computed expert trajectories.
Specifically, we first employ a `warm-up' phase with $L_n$ for 1 epoch to achieve initial convergence, followed by 9 epochs training with dual-track collaborative learning.
Then we randomly select starting points from pre-computed expert trajectories to initialize student models for dataset distillation.
Here, we follow \cite{vl_distill,lors} to synthesize text embeddings directly instead of full captions. More implementation details, such as learning rates and trajectory length, are provided in Appendix \ref{sec: hyperparameter} due to the page limit.

\subsection{Comparisons with State-Of-The-Art (SOTA)}
We compare our MDW against SOTA multi-modal dataset distillation methods to highlight its efficacy, including two main categories: (1) Coreset selection methods such as K-center \cite{sener2018active} and Forgetting \cite{toneva2018empirical}; (2) Dataset Distillation methods including LoRS \cite{lors} and VL-Distill \cite{vl_distill}.
Moreover, we report the results of LoRS on clean Flickr30K and COCO by discarding the PMPs, namely LoRS-C. 
We further enhance LoRS with existing PMP-robust methods \cite{huang2021learning} to filter out PMPs, denoted as LoRS+.
Clearly, LoRS-C and LoRS+ are strong baselines for noisy multi-modal dataset distillation since they effectively address the challenging PMPs, providing a solid foundation for evaluating our MDW.

\noindent\textbf{Flickr30K and COCO with Synthetic Noise.} 
As Flickr30K and COCO are originally well-annotated through meticulous human verification, we inject PMPs by randomly shuffling images for specific noise ratios $\eta$ (\ie, 0\%, 30\%, 50\%) to simulate synthetic noise.
\Cref{table:flicker} details the results of distilling such noisy data into 100 samples, accounting for less than 0.1\% of the original dataset size.

Specifically, coreset selection methods fail as they rely on original training data, producing subsets with inadequate informativeness and diversity, particularly in complex multi-modal datasets.
Notably, although MDW is designed for noisy multi-modal dataset distillation, our fine-grained correspondence-enhanced distillation strategy effectively guides the distillation process to better capture and refine multi-modal correspondence patterns within distilled data, thereby surpassing previous strong competitor LoRS by over 10\% in R\_sum even in noise-free cases.

When the data is contaminated by PMPs, our MDW substantially outperforms previous SOTAs by over 20\% R\_sum, with the performance gap widening as the noise ratio increases. This trend highlights previous methods' vulnerability to PMPs and the necessity for robust multi-modal dataset distillation techniques.
Moreover, although filtration-based methods, LoRS-C and LoRS+, mitigate the impact of PMPs, they struggle with severe information loss from real datasets. 
Conversely, our MDW proposes dual track collaborative learning to effectively mine reliable supervision from negative matches with certifiable noise tolerance, greatly enhancing distillation efficacy.

\begin{wraptable}{r}{0.6\textwidth}
    \vspace{-5mm}
	\caption{
    Distilling real-world noise CC104K into 100 samples. Our MDW and PMP-robust LoRS variants are highlighted in gray.
	Model on original dataset training achieves R@1/5/10 of 12.9/33.4/45.4 (Image) 12.0/32.2/44.2 (Text).
    }
  \vspace{-3mm}
	\makeatletter\def\@captype{table}
	\resizebox{\linewidth}{!}{
		\begin{tabular}{c|ccc|ccc|c}
			\toprule[1.5pt]
			\multirow{2}{*}{Methods}& \multicolumn{3}{c|}{Image$\longrightarrow$Text}&\multicolumn{3}{c|}{Text$\longrightarrow$Image}&\multirow{2}{*}{R\_Sum}\\
			\cline{2-7}
			&R@1&R@5&R@10&R@1&R@5&R@10&\\
			\hline
			K-center& 0.8 & 3.1 & 4.6 & 0.6 & 2.2 & 3.1 & 14.4 \\
			Forget& 0.7 & 2.9 & 5.1 & 0.5 & 2.2 & 3.8 & 15.2 \\
			vl-distill& 2.7 & 9.3 & 14.9 & 1.7 & 6.0 & 10.3 & 44.9 \\
			LoRS& 4.3 & 13.9 & 21.4 & 2.9 & 9.8 & 15.6 & 67.9 \\
			\rowcolor{tabhighlight}LoRS+& 5.0 & 15.3 & 23.4 & 3.0 & 11.2 & 17.3 & 75.2 \\
			\rowcolor{tabhighlight}Ours& \textbf{5.9} & \textbf{16.9} & \textbf{26.0} & \textbf{4.0} & \textbf{13.2} & \textbf{21.1} & \textbf{87.1} \\
			\bottomrule[1.5pt]
		\end{tabular}
	}
  \label{cc104k}
  \vspace{-3mm}
\end{wraptable}

\noindent\textbf{CC104K with Real-World Noise.}
The CC104K, automatically collected from the Internet, naturally contains approximately 30\% PMPs. To evaluate our MDW's performance in realistic noisy scenarios, we distilled CC104K into 100 sample pairs without introducing any additional noise.
\Cref{cc104k} highlights that MDW consistently surpasses all baselines across all metrics.
Notably, MDW exceeds the current SOTAs, \ie, LoRS, by around 30\% in R\_sum, highlighting its appealing practicality.

\subsection{How PMPs Affect Dataset Distillation?}
\begin{figure}[!t]
    \centering
    \begin{subfigure}[b]{0.49\textwidth}
      \centering
      \includegraphics[width=\textwidth]{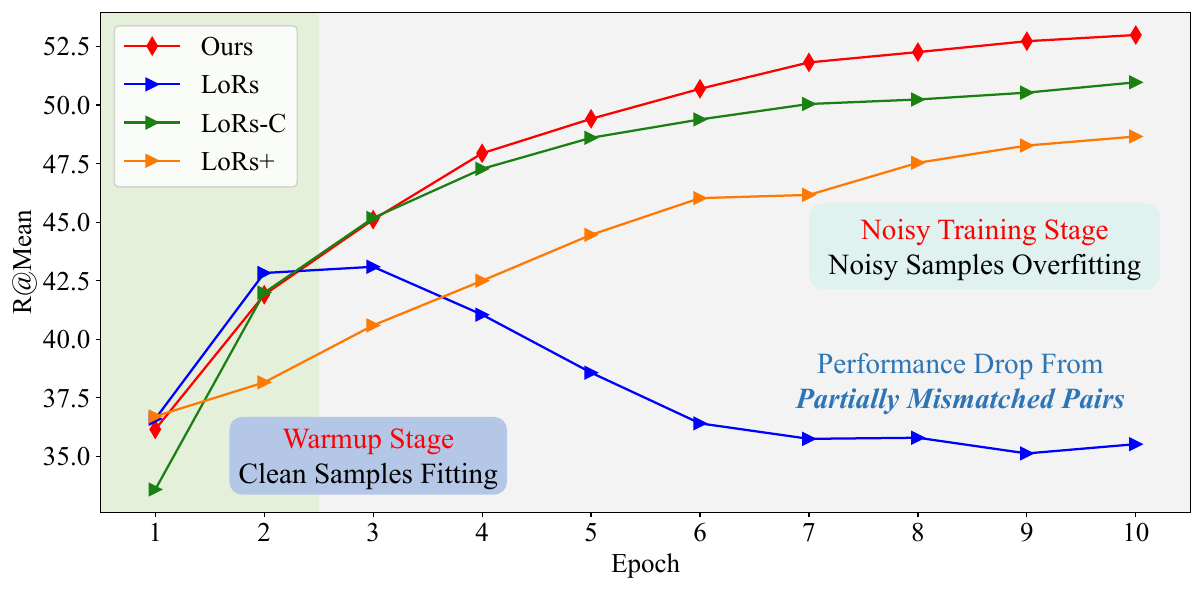}
      \vspace{-6mm}
      \caption{Per epoch performance on 30\% noise Flickr30K.}
      \label{fig:trajectory}
    \end{subfigure}
    \hfill
    \begin{subfigure}[b]{0.49\textwidth}
      \centering
      \includegraphics[width=\textwidth]{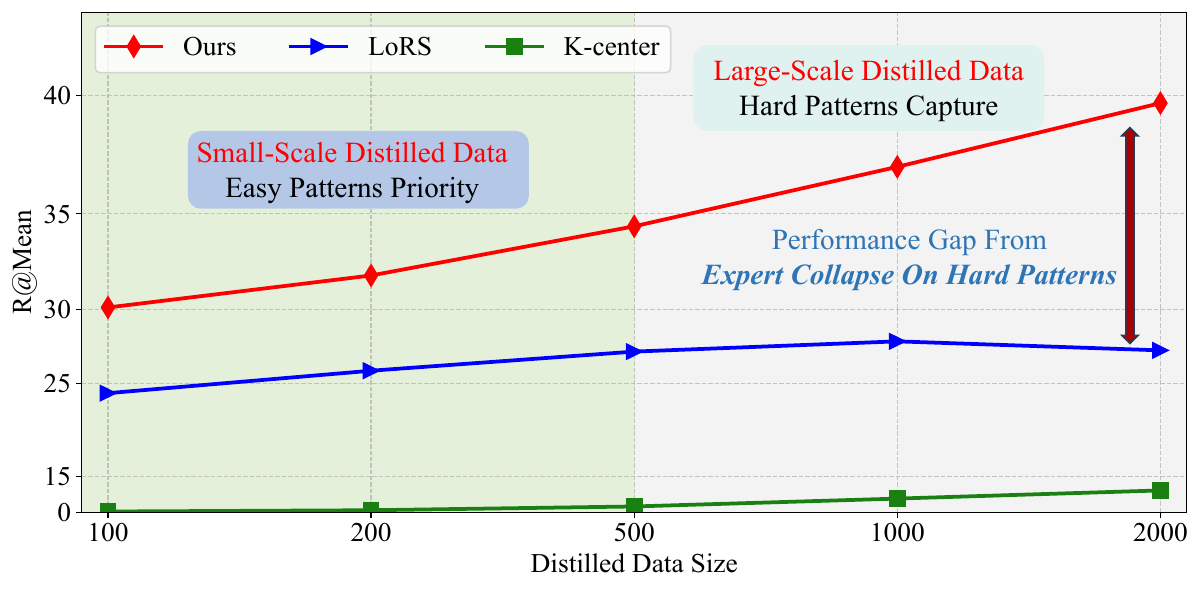}
      \vspace{-6mm}
      \caption{Scalability analysis on 30\% noise Flickr30K.}
      \label{fig:scalability}
    \end{subfigure}
    \vspace{-1mm}
    \caption{Analysis of the PMP-robustness and scalability among dataset distillation algorithms.}
    \label{fig:analysis}
    \vspace{-6mm}
  \end{figure}

We attribute the failure of noisy multi-modal dataset distillation to PMPs that mislead expert training and lead to erroneous model trajectory. To validate this hypothesis and our MDW's efficacy, \Cref{fig:trajectory} shows expert trajectories under 30\% noise Flickr30K.
As trajectories lie in complex model parameter space, we use the per-epoch test performance of expert models as a proxy for the training trajectory.

During the first two epochs, the memorization effect enables the model to prioritize clean samples, rapidly improving performance and laying the foundation for noise sample filtration \cite{huang2021learning}.
As training progresses, the vanilla expert model gradually overfits to noisy samples with a performance drop, while LoRS-C and LoRS+ steadily enhance performance by effective PMP filtration. 
However, such filtration-based methods inevitably cause information loss in the real dataset.
Conversely, our MDW effectively leverages negative matches with certifiable noise tolerance to enhance expert efficacy.

\subsection{Ablation Study}
This section conducts ablation studies on our MDW to investigate its efficacy. More experimental results are available in the Appendix, including efficiency analyses and real-world PMPs visualization.

\noindent\textbf{Practicality and Scalability.}
\Cref{table: cc_200} presents the results of distilling real-world noise CC104K into 200 and 500 pairs, where the larger distilled datasets capture more knowledge from real datasets and enhance performance across all methods.
Furthermore, we expand the distilled data size on the 30\% noise Flickr30K to comprehensively evaluate our MDW's practicality and scalability. 
As shown in \Cref{fig:scalability}, LoRS's performance improves slowly with increasing distilled samples and even degrades when using 2K samples.
This is because, as the distilled dataset size grows, the target multi-modal correspondence patterns shift from simple to more complex \cite{guotowards}.
However, due to its noise sensitivity shown in \Cref{fig:trajectory}, LoRS tends to overfit to PMPs during the late training stage when learning harder patterns \cite{arpit2017closer}.
Additionally, LoRS fails to effectively capture and refine simple patterns, as its optimization treats all pixels equally without focus on correspondence-discriminative regions.

\begin{wraptable}{r}{0.5\textwidth}
\centering
\vspace{-6mm}
    \caption{Component analyses on CC104K.}
    \vspace{-2mm}
    \label{table:ablation}
    \setlength{\tabcolsep}{2.1pt}
        \begin{tabular}{cccc|c|c}
            \hline
            \multicolumn{4}{c|}{Methods}& \multicolumn{1}{c|}{Image$\longrightarrow$Text}&\multicolumn{1}{c}{Text$\longrightarrow$Image}\\
            \hline
            $L_{mae}$& $L_n$ &$\tilde{\mathcal{P}}$&CEO&R@1/5/10&R@1/5/10\\
            \bottomrule[0.5pt]
            &  & & &2.7/9.3/14.9&1.7/6.0/10.3\\
            \checkmark&& &&4.7/14.0/21.7&2.6/8.2/12.0\\
            &\checkmark& &&4.9/14.9/22.8&2.7/9.1/14.1\\
            &\checkmark&\checkmark&&5.2/15.7/23.8&3.4/11.5/18.7\\
            &\checkmark&\checkmark&\checkmark&\textbf{5.9}/\textbf{16.9}/\textbf{26.0}&\textbf{4.0}/\textbf{13.2}/\textbf{21.1}\\
            \hline
        \end{tabular}
        \vspace{1mm}
	\centering
	\newcommand{\tabincell}[2]{\begin{tabular}{@{}#1@{}}#2\end{tabular}}
	\centering
	\caption{Distilled data size ablation on CC104K. 
	Model trained on full data achieves R@1/5/10 of 12.9/33.4/45.4 (Image) and 12.0/32.2/44.2 (Text).}
	\vspace{-2mm}
	\label{table: cc_200}
	\resizebox{\linewidth}{!}{ 
		\begin{tabular}{c|c|ccc|ccc}
		\toprule[1.5pt]
		\multirow{2}{*}{Pairs}&\multirow{2}{*}{Methods}&\multicolumn{3}{c|}{Image$\longrightarrow$Text}&\multicolumn{3}{c}{Text$\longrightarrow$Image}\\
		\cline{3-8}
			&&R@1&R@5&R@10&R@1&R@5&R@10\\
			\midrule
			\multirow{6}{*}{200} &K-center& 1.5 & 5.7 & 9.3 & 1.0 & 3.3 & 5.7  \\
			&Forget& 1.4 & 5.1 & 8.5 & 0.7 & 3.1 & 5.0 \\
			&vl-distill& 2.9 & 9.6 & 15.2 & 2.1 & 7.3 & 11.5  \\
			&LoRS& 4.5 & 15.8 & 23.6 & 3.0 & 10.5 & 16.5  \\
			&\cellcolor{tabhighlight}LoRS+&\cellcolor{tabhighlight}5.5 &\cellcolor{tabhighlight}17.9 &\cellcolor{tabhighlight}27.5 &\cellcolor{tabhighlight}4.0 &\cellcolor{tabhighlight}13.1 &\cellcolor{tabhighlight}20.3  \\
			&\cellcolor{tabhighlight}Ours&\cellcolor{tabhighlight}\textbf{6.4} & \cellcolor{tabhighlight}\textbf{19.0} &\cellcolor{tabhighlight}\textbf{29.1} &\cellcolor{tabhighlight}\textbf{4.7} &\cellcolor{tabhighlight}\textbf{15.2} &\cellcolor{tabhighlight}\textbf{22.8}  \\
			\midrule
            \multirow{6}{*}{500} &K-center& 2.2 & 7.8 & 12.4 & 1.4 & 5.6 & 9.1 \\
			&Forget& 1.9 & 7.1 & 11.3 & 1.0 & 4.0 & 7.6  \\
			&vl-distill& 3.1 & 11.5 & 17.9 & 2.5 & 8.3 & 14.4 \\
			&LoRS& 5.1 & 16.5 & 25.1 & 4.1 & 12.7 & 20.8 \\
			&\cellcolor{tabhighlight}LoRS+&\cellcolor{tabhighlight}6.6 &\cellcolor{tabhighlight}19.7 &\cellcolor{tabhighlight}30.3 &\cellcolor{tabhighlight}5.2 &\cellcolor{tabhighlight}16.4 &\cellcolor{tabhighlight}25.3  \\
			&\cellcolor{tabhighlight}Ours&\cellcolor{tabhighlight}\textbf{7.4} & \cellcolor{tabhighlight}\textbf{21.9} &\cellcolor{tabhighlight}\textbf{31.8} &\cellcolor{tabhighlight}\textbf{5.4} &\cellcolor{tabhighlight}\textbf{17.7} &\cellcolor{tabhighlight}\textbf{25.9}  \\
			\bottomrule[1.5pt]
	\end{tabular}}
	\vspace{-7mm}
\end{wraptable}
Conversely, our MDW shows appealing scalability, attributed to its fine-grained correspondence-enhanced distillation and certified noise robustness. Remarkably, with only 2K distilled samples (less than 1\% of the original dataset), MDW achieves a 40 mean recall, equivalent to around ~74\% of the full-data performance.
More results on Flickr30K and COCO are in Appendix \ref{sec: sample_size}.

\noindent\textbf{Component Analyses.}
\Cref{table:ablation} ablates our MDW's key components, such as dual-track collaborative learning ($L_{mae},L_n$), soft matching probability $\tilde{\mathcal{P}}$ and corresponding-enhanced optimization (CEO).
Without all proposed components, MDW reduces to the vanilla multi-modal dataset distillation baseline, vl-distill \cite{vl_distill}, with severe vulnerability to PMPs.
Notably, by considering informative hard negative pairs, $L_n$ consistently improves all metrics.
In detail, $L_n$ greatly enhances expert model to mine effective multi-modal priors for dataset distillation, revealing the necessity for alleviating real data's information loss and validating our theoretical analyses.
Moreover, our soft matching probability $\tilde{\mathcal{P}}$ captures fine-grained sample-level correspondences to enrich distilled dataset's information density.
Additionally, our CEO boosts R\_sum by 10\%, highlighting the necessity of focusing on correspondence-discriminative regions during distillation.
The optimal performance is attained with all components, underscoring the essential role of each in noisy multi-modal dataset distillation.

\begin{wraptable}{r}{0.6\textwidth}
    \centering
    \vspace{-5mm}
    \caption{Cross-architecture generalization on 30\% noise Flickr30K. The data is synthesized with NFNet+BERT at a scale of 100 and evaluated on various architectures.}
    \vspace{-2mm}
    \label{tab: cross architecture}
    \resizebox{\linewidth}{!}{
    \begin{tabular}{cc|c|c}
    \hline
    \multirow{2}{*}{Methods}&\multirow{2}{*}{Model}&Image$\longrightarrow$Text&Text$\longrightarrow$Image\\
     &   & R@1/5/10 & R@1/5/10 \\
    \hline
    \multirow{3}{*}{LoRS}   & NFNet+BERT & 10.9/31.5/43.5 & 7.0/20.6/31.9 \\
    & ResNet+BERT& 0.8/3.3/6.0 & 0.5/2.1/4.0 \\
    & RegNet+BERT& 1.1/4.4/7.4 & 0.7/3.2/5.5 \\
    \hline
    \multirow{3}{*}{Ours}   & NFNet+BERT &14.3/39.1/53.7  & 8.6/26.2/38.7  \\
    & ResNet+BERT& 3.9/12.1/19.0 & 1.7/6.3/10.6 \\
    & RegNet+BERT &3.7/12.1/18.9& 1.5/6.4/10.7 \\
    \hline
    \end{tabular}
    }
    \vspace{-3mm}
  \end{wraptable}

\noindent\textbf{Cross-Architecture Generalization.}
We assess the distilled dataset's generalizability with a cross-architecture setting. We distill the dataset with NFNet+BERT and evaluate it with other architectures, \eg, RegNet \cite{radosavovic2020designing} and ResNet \cite{he2016deep}. Following \cite{lors}, we do not assess the text encoder's generalizability, as it remains frozen during training and distillation. 
\Cref{tab: cross architecture} shows that the strong competitor LoRS inevitably captures noise from the real dataset during dataset distillation, severely compromising its generalizability. Conversely, our MDW enhances robustness against PMPs and enriches the distilled dataset's information density with fine-grained correspondence-enhanced distillation. 
Therefore, MDW effectively captures generalizable knowledge, showcasing superior cross-architecture generalizability. 
Moreover, our MDW's performance drop is primarily due to the inherent limitations of architectures themselves, \eg, ResNet or RegNet trained on the full dataset achieve IR@1=17\% and TR@1=24\%, whereas NFNet achieves IR@1=25\% and TR@1=33\%.

\begin{figure}[!t]
  \centering
  \includegraphics[width=1\textwidth]{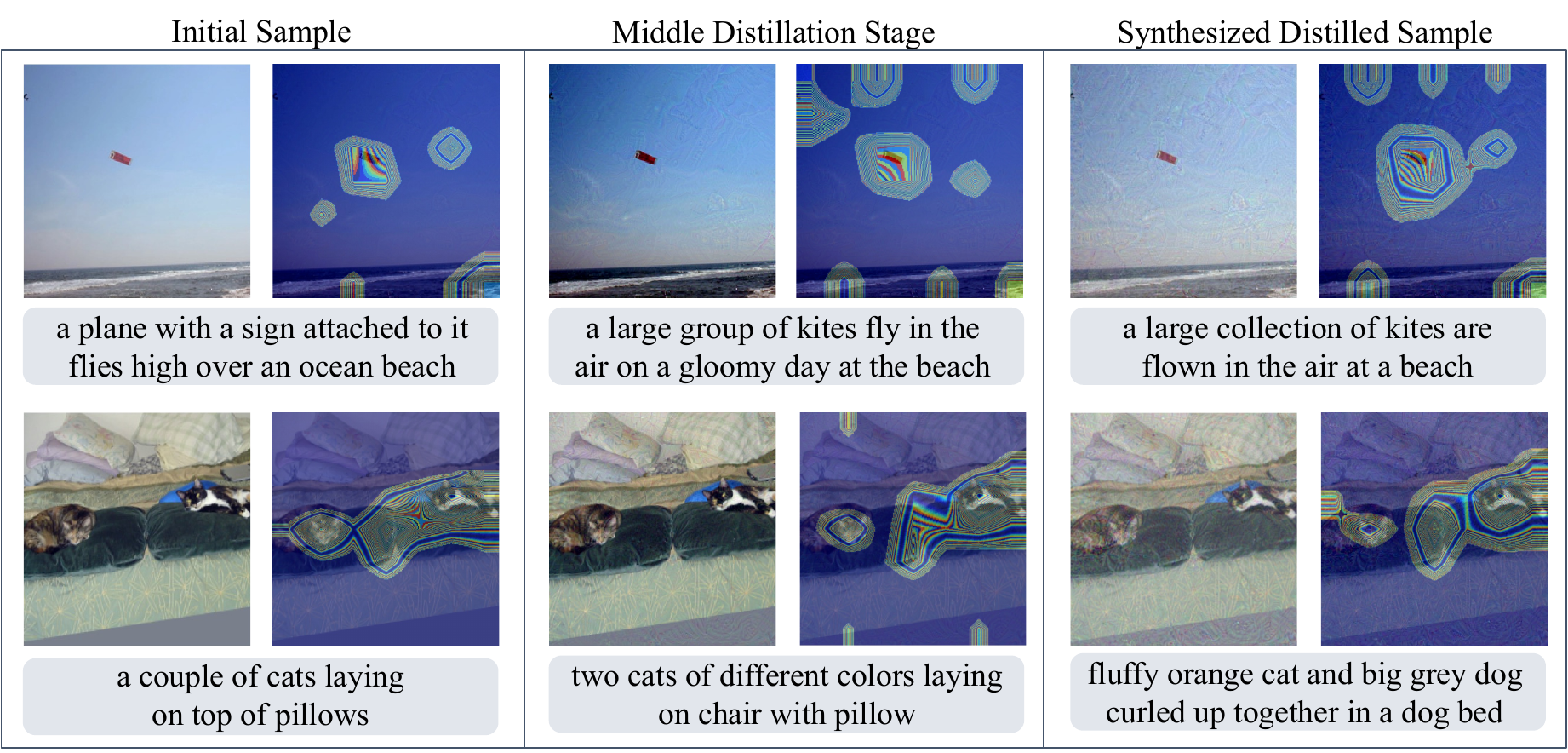}
  \vspace{-5mm}
  \caption{Illustrations of synthetic distilled samples with correspondence-discriminative regions.}
  \vspace{-3mm}
  \label{fig:distilled_sample}
\end{figure}

\begin{wraptable}{r}{0.5\textwidth}
	\vspace{-5mm}
	\caption{Architecture ablation on real-world noise CC104K with 100 sample pairs and R\_sum metric.}
	\vspace{-2mm}
      \label{tab: architecture}
      \resizebox{\linewidth}{!}{
      \begin{tabular}{cc|cccc}
      \hline
      Visual&Textual& vl-distill & LoRS  & Ours\\
      \hline
      NF\_ResNet   & BERT & 7.4 & 9.6 & 11.0\\
      NF\_RegNet & BERT & 7.7 & 9.9 & 11.9\\
      VIT & BERT& 6.1 & 7.3& 8.9\\
      NFNet   & DistillBert & 9.2 & 12.9  & 15.4 \\
      NFNet & CLIP& 24.4 & 29.4  & 33.1 \\
      \hline
      \end{tabular}}
      \vspace{-4mm}
\end{wraptable}

\noindent\textbf{Ablation on Model Architecture.}
To evaluate our MDW's adaptability to different modality encoders, we extend experiments on real-world noise CC104K with various image and text encoders, \eg,  ResNet \cite{he2016deep}, RegNet \cite{radosavovic2020designing} and VIT \cite{dosovitskiy2020image} for visual, DistilBERT \cite{sanh2019distilbert} and CLIP \cite{radford2021learning} for textual. 
Owing to its fine-grained correspondence-enhanced distillation and certifiable noise tolerance to risky data noise, \Cref{tab: architecture} highlights that our MDW consistently surpasses all competitors across various architectures by a notable margin, confirming its appealing generalizability.

\noindent\textbf{Distilled Data Visualization.}
\Cref{fig:distilled_sample} shows synthetic distilled image-text pairs on 30\% noise COCO. Following \cite{vl_distill}, the texts are retrieved as the nearest sentences in the training set to the distilled text embeddings.  
Notably, the distilled images incorporate high-frequency details to enhance generalization performance \cite{wang2020high}, exhibiting a DeepDream-Style \cite{zeiler2014visualizing} typical in dataset distillation. Correspondingly, the distilled texts are significantly adjusted to align with these images.
Moreover, \Cref{fig:distilled_sample} further verifies the efficacy of our correspondence-enhanced optimization, where correspondence-discriminative regions are accurately identified to receive more updates. 
Therefore, MDW effectively captures and refines multi-modal correspondence patterns within the distilled data, \eg, kite patterns are progressively encoded into the sky region in the top row of \Cref{fig:distilled_sample}, greatly improving the distilled data's information density and efficacy.
More results are in Appendix \ref{sec: more_distilled_sample_vis}.  

\begin{figure*}[!t]
    \centering
    \includegraphics[width=\linewidth]{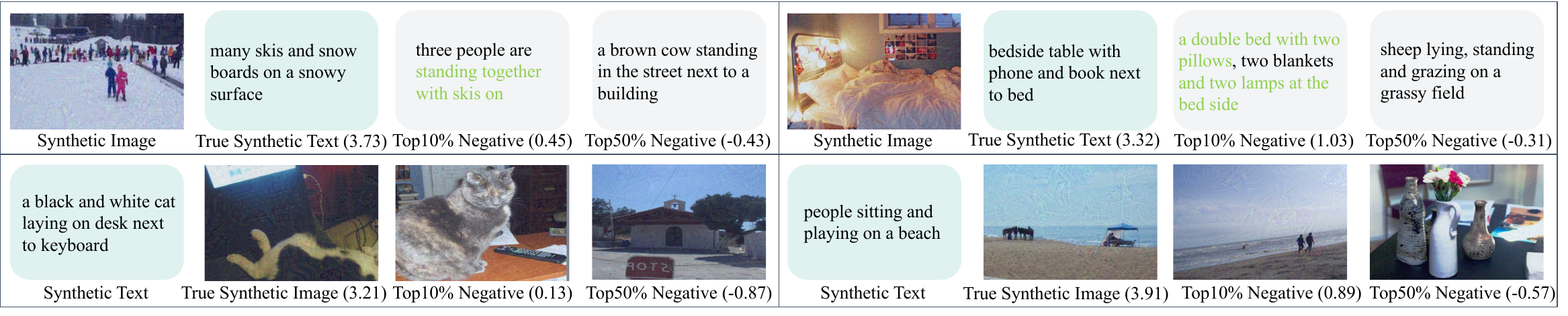}
    \vspace{-5mm}
  
  \caption{
    Synthetic matching probability visualization on distilling 30\% noise COCO dataset.
  }
  \vspace{-6mm}
  \label{fig: soft_prob}
  \end{figure*}

\noindent\textbf{Soft Matching Probability Visualization.}
\Cref{fig: soft_prob} shows the learned matching probability logits on 30\% noise COCO, including paired synthetic distilled data and their top 10\% and 50\% negatives. Notably, MDW ensures that paired synthetic samples dominate the matching process with the highest logits while identifying informative negatives among distilled data.
For example, high matching logits are assigned to negatives with human-perceived similarities, \ie, shared beach or cat, while effectively suppressing correspondences among regular negatives. These results validate MDW's efficacy in capturing fine-grained correspondences to enrich distilled data's information density.

\section{Conclusion}

This paper presents MDW, the first work for noisy multi-modal dataset distillation. 
MDW employs fine-grained correspondence-enhanced distillation strategy to enhance distilled data's information density and efficacy.
Moreover, MDW introduces dual-track collaborative learning to capture multi-modal priors from noisy real data with certifiable noise tolerance, effectively alleviating information loss from the real data. Extensive experiments across benchmarks validate MDW's superior empirical and theoretical efficacy with remarkable scalability, surpassing existing methods by over 15\%.

\section*{Broader Impacts}
Our MDW significantly reduces the data storage and computation burden of multi-modal model training, greatly advancing the pursuit of green AI.
Additionally, addressing partially mismatched pairs further enhances multi-modal dataset distillation by minimizing manual data verification, improving data utilization, and boosting the robustness of multi-modal systems in noisy real-world scenarios, \etc 
Moreover, our MDW's generalizability across various architectures and noise ratios highlights its potential for widespread adoption in multi-modal learning research and applications.

\bibliographystyle{plain}
\bibliography{main}

\newpage
\appendix

\begin{figure*}[!t]
  \centering
  \subfloat[Initial Distribution]{{\includegraphics[width=0.24\textwidth]{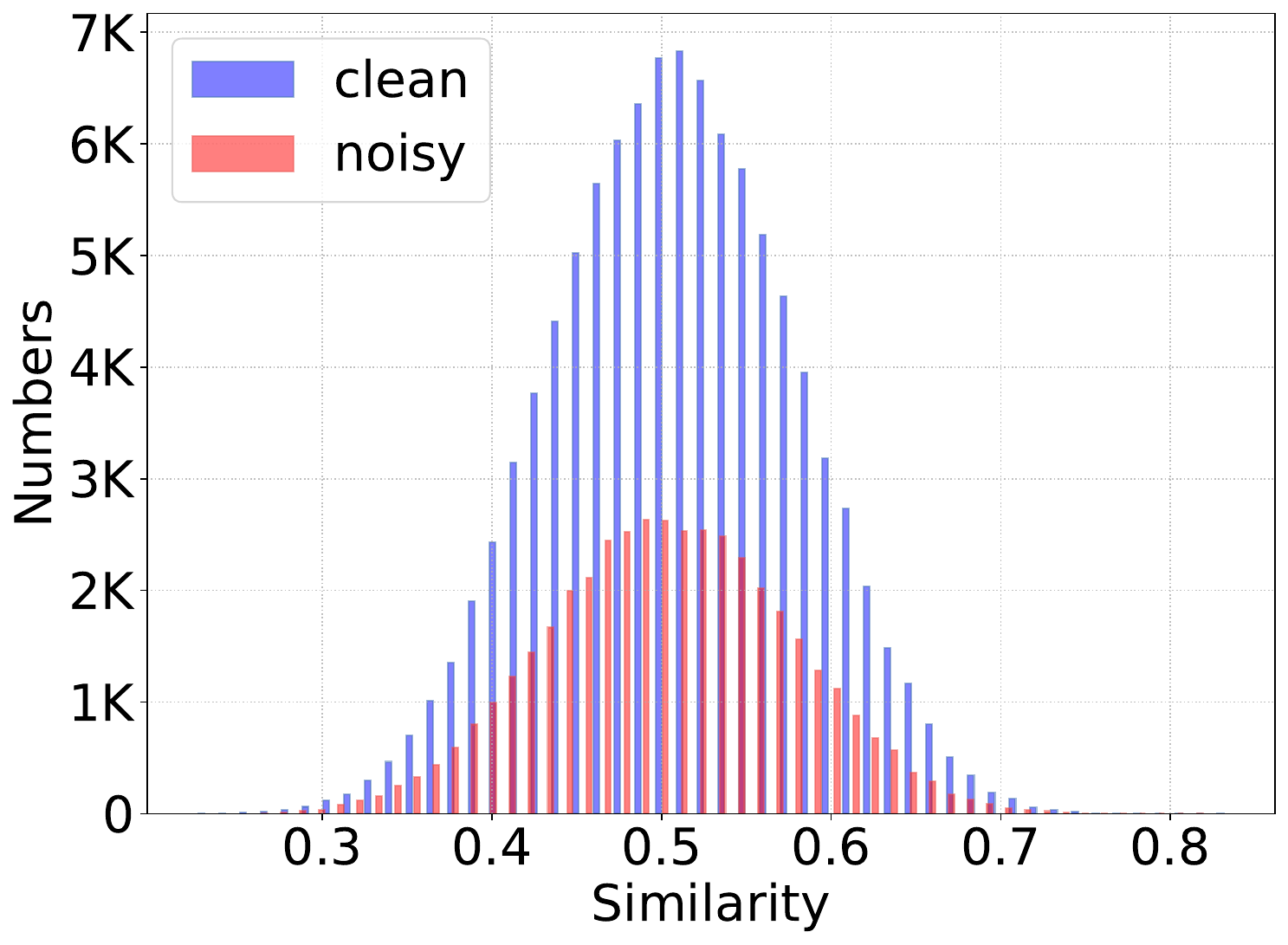}}} \hfill
  \subfloat[Epoch 1 (Warmup)]{{\includegraphics[width=0.24\linewidth]{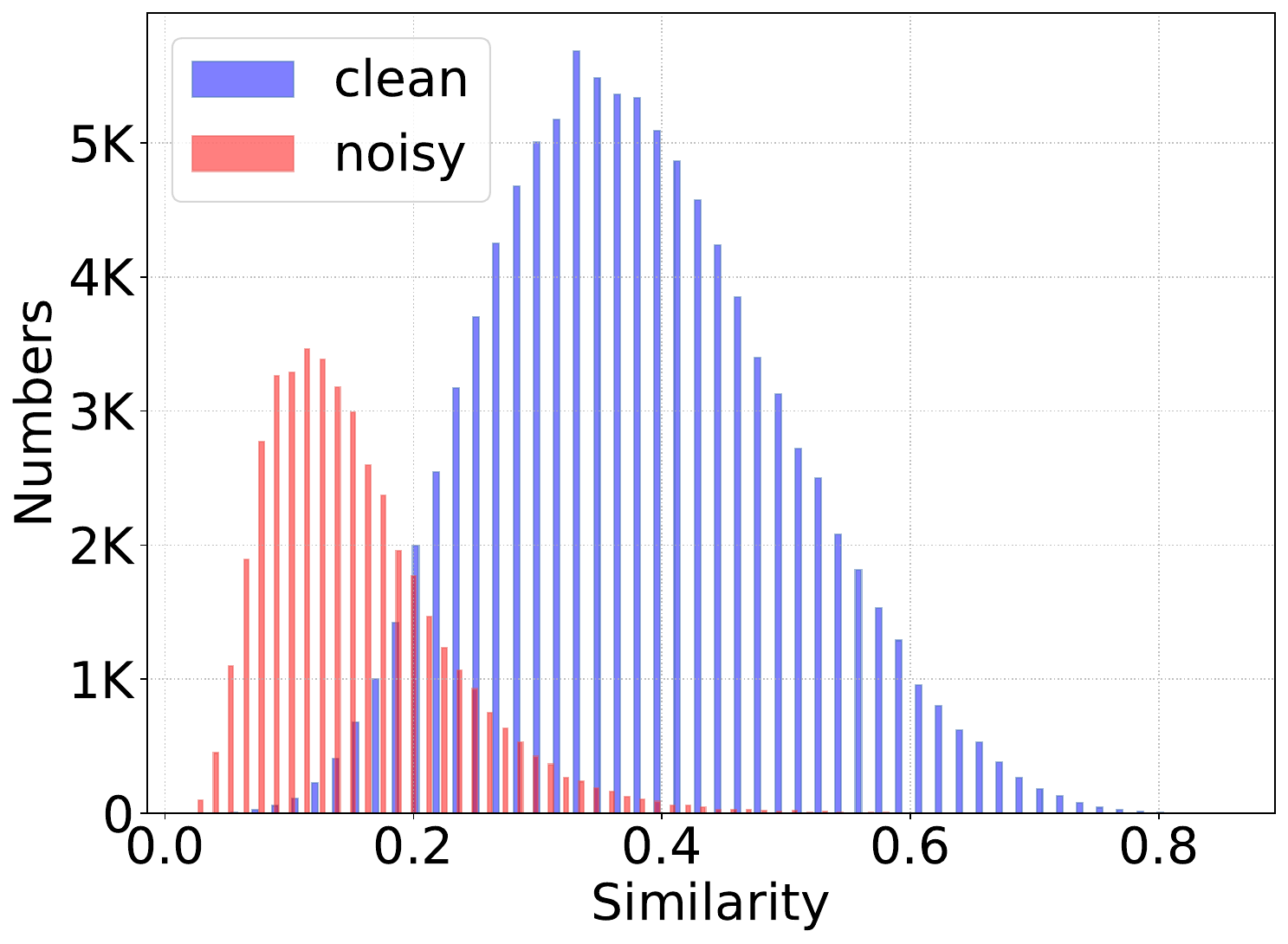}}} \hfill
  \subfloat[Epoch 5 (Middle)]{{\includegraphics[width=0.24\linewidth]{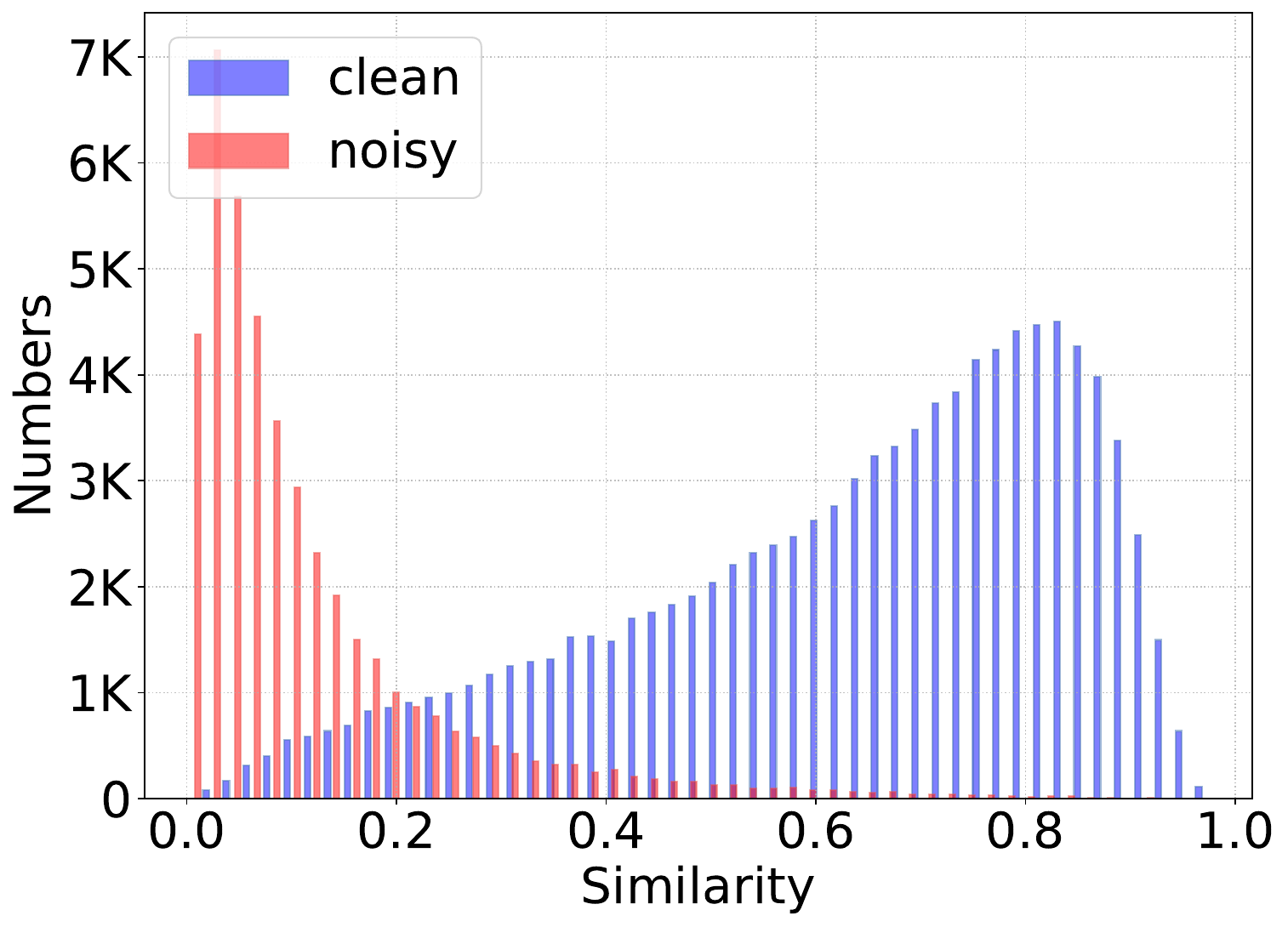}}} \hfill
  \subfloat[Epoch 10 (Last)]{{\includegraphics[width=0.24\linewidth]{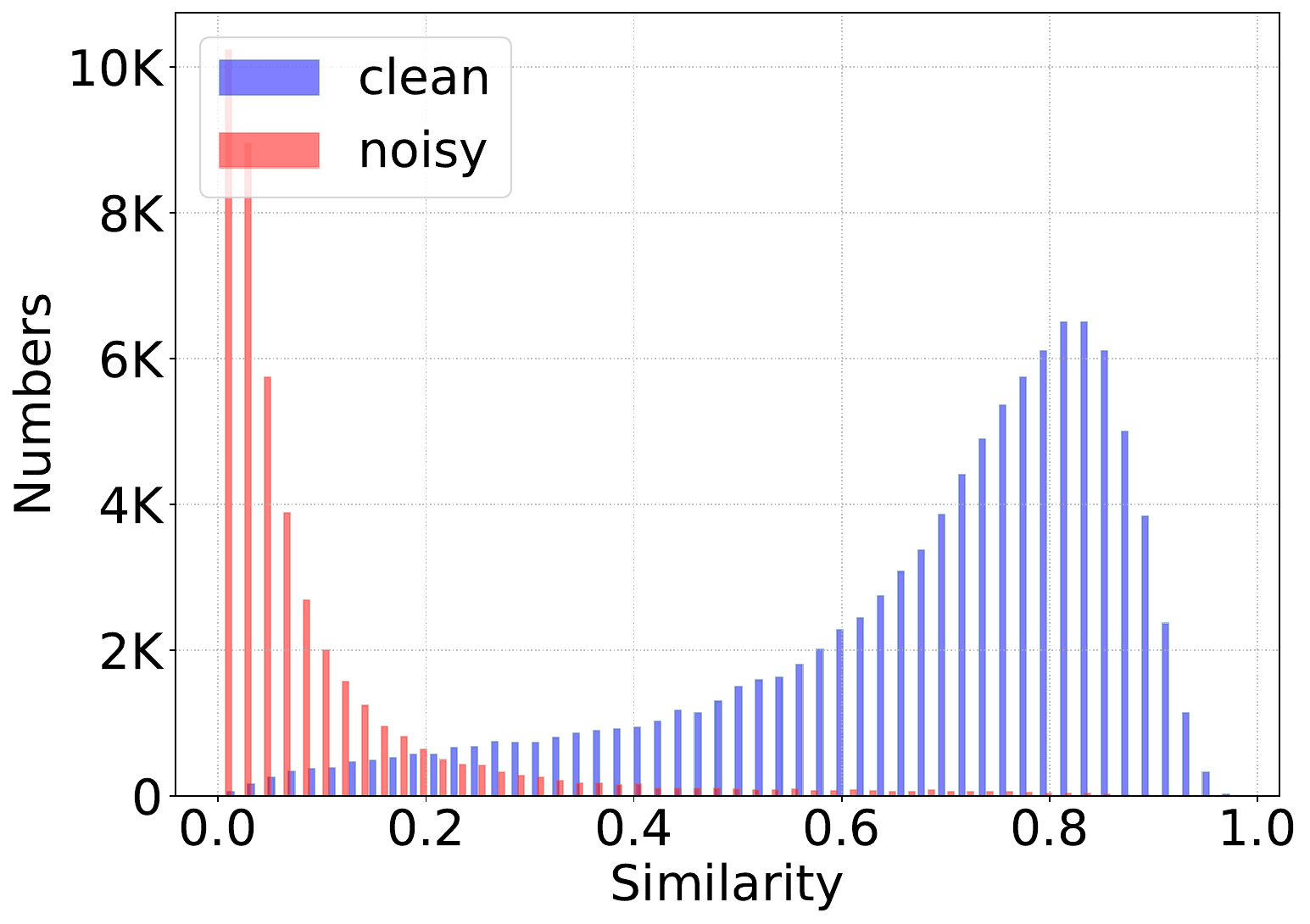}}}
  \vspace{-2mm}
  \caption{Visualization of our MDW's global similarity distribution at different training stages on Flickr30K with 30\% noise. Thanks to MDW, clean pairs' similarity gradually shifts to the right (high) while noisy ones tightly gather to the left (low).
  }
  \vspace{-4mm}
  \label{fig: sim distribution}
\end{figure*}

\begin{figure*}[!t]
  \centering
  \includegraphics[width=\linewidth]{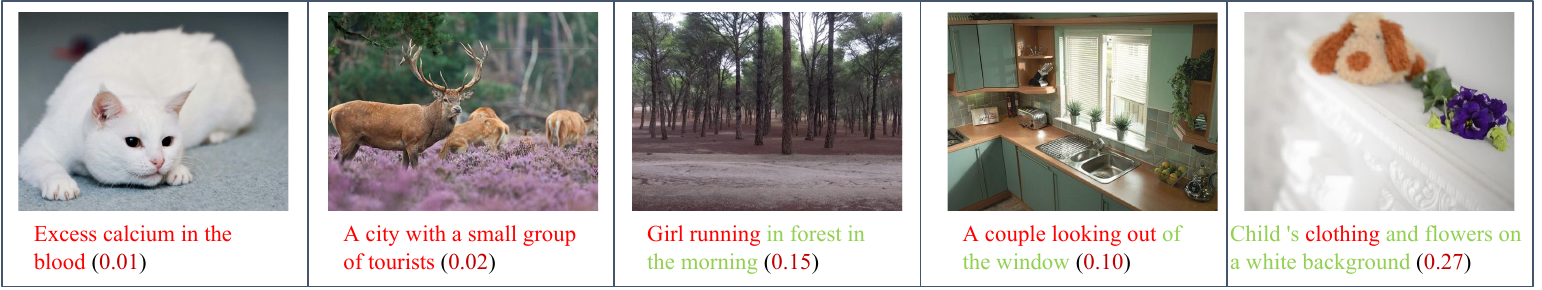}
  \vspace{-4mm}
  \caption{
    Illustration of real-world PMPs detected by our concensus-driven sample selection on CC104K dataset with real-world noise. We show corresponding similarity predictions in brackets and highlight the matched words in green while the mismatched ones in red.
    }
  \vspace{-4mm}
    \label{fig: detected PMPs}
  \end{figure*}

\section{PMP Filtration and Visualization}
\label{sec: PMP_filtration}
\subsection{PMP Filtration with Neural Networks' Memorization Effect}
To address the risky PMPs in real datasets $\mathcal{D}_\eta$, we follow previous works \cite{huang2021learning,yang2024robust,zhao2024mitigating,dang2024noisy,yang2023bicro,dang2025disentangled} to explore neural networks' memorization effect to devise sample selection strategy for accurately identifying PMPs, preventing them from disrupting expert training.
Empirically, the memorization effect of neural networks \cite{arpit2017closer,yao2020searching} reveals that neural networks tend to learn clean samples first and then gradually fit noisy samples during training.
In image-text matching, we observe that clean samples will exhibit higher similarities than noisy ones after the initial few epochs.
Therefore, we explore such difference between clean and noisy samples from both local and global perspectives for accurate PMP identification, namely \textbf{consensus-driven sample selection}.

\noindent\textbf{Global-Level Partition.}
We follow \cite{han2023noisy} to compute the per-sample similarity score set as $\mathcal{S} = \{s_i=h(V_i,T_i)|(V_i,T_i)\in \mathcal{D}_\eta\}$.
Then we estimate the global similarity distribution of all training samples by using Expectation-Maximization algorithm \cite{moon1996expectation} to fit $\mathcal{S}$ with a two-component Beta Mixture Model (BMM) \cite{arazo2019unsupervised}, \ie, $p(s_i)  = \sum_{k = 1}^{K} \lambda_k \phi(s_i \mid \alpha_k, \beta_k)$.
Here, $K=2$ denotes the clean and noisy component of similarity distribution; $\lambda_k$ is the mixture coefficient; $\phi(s_i \mid \alpha_k, \beta_k)$ is the beta probability density function of $k$-th component with parameters $\alpha_k, \beta_k>0$.
Given the neural networks' memorization effect, we identify the component with a higher mean (\ie, higher similarity) in the fitted BMM as the clean component. The posterior probability $p_i = p(k|s_i) = \frac{p(k)p(s_i|\alpha_k,\beta_k)}{p(s_i)}$ of each sample $(V_i,T_i)$ belonging to clean component $k$ is then computed.
Finally, the noisy dataset $\mathcal{D}_\eta$ is filtered by setting a threshold $\delta$ on the posterior probability $p_i$, \ie,
\begin{equation}
    \mathcal{D}_{\text{clean}}^{\text{global}} = \{(V_i,T_i)|p_i>\delta, \forall (V_i, T_i)\in\mathcal{D}_\eta\}.
\end{equation}

Note that the overconfidence of global similarity distribution modeling on a single threshold \( \delta \), as shown in \Cref{fig: sim distribution}, may misjudge noisy samples near the threshold boundary. To address this issue, we further introduce a local-level similarity difference to further refine sample selection.

\noindent\textbf{Local-Level Partition.} 
Given a batch of multi-modal inputs $(V_i, T_j)_{i,j=1}^B$ with batch size $B$, neural networks' memorization effect drives the model to primarily learn clean samples during early training, leading to a \textbf{\textit{unimodal}} similarity distribution at $i$-th sample.
In contrast, it struggles with mismatched pairs, failing to distinguish them from their negatives and thus exhibiting a more \textbf{\textit{uniform}} similarity distribution.
Therefore, we refine the sample selection with local similarity differences as follows:
\begin{equation*}
    \mathcal{D}_{\text{clean}}^{\text{local}} = \{(V_i,T_i)|i=\arg\max\nolimits_jh_{ij}\ \wedge \ i=\arg\max\nolimits_jh_{ji}\},
\end{equation*}
where $i,j<B$ and $\arg\max$ ensures the maximum similarity aligns with the diagonal within each batch, \ie, potential clean samples in multi-modal learning.

The final clean subset is selected by the consensus of both partitions to ensure clean subset's purity, \ie, 
\begin{equation}
  \mathcal{D}_{\text{clean}}= \mathcal{D}_{\text{clean}}^{\text{global}}\cap\mathcal{D}_{\text{clean}}^{\text{local}}
  \label{eq: clean_data}
\end{equation}
Moreover, unlike existing metrics, \eg, energy scores \cite{liu2020energy}, with auxiliary training, our consensus-driven sample selection strategy is training-free and parameter-independent. This design prevents redundant gradients from compromising the efficacy of trajectory matching.

\subsection{Visualization on sample selection}
\Cref{fig: sim distribution} visualizes the global similarity distribution during training. After warmup, the initially uniform distribution evolves into a distinct bimodal pattern, with clean samples exhibiting higher similarity than noisy ones. 
This validates neural network's memorization effect, underpinning our consensus-driven sample selection strategy.
As training progresses, the similarity of clean samples steadily increases, while that of noisy samples decreases, leading to a more pronounced bimodal distribution and highlighting our consensus-driven sample selection's efficacy in identifying PMPs.

  \subsection{Visualization of PMPs}
  \Cref{fig: detected PMPs} shows real-world PMPs detected on CC104K. Beyond detecting obvious PMPs with completely irrelevant information, our consensus-driven sample selection effectively captures subtle semantic misalignments within input pairs, \eg, the omission of key details like ``girl" and ``couple" in the foreground or the textual misclassification of ``child toy" as ``child clothing".
  These findings highlight our consensus-driven sample selection's impressive efficacy and real-world applicability.

\begin{table*}[!t]
	\newcommand{\tabincell}[2]{\begin{tabular}{@{}#1@{}}#2\end{tabular}}
	\centering
	\caption{Distilling noisy Flickr30K and COCO into 200/500 samples. Our MDW and PMP-robust LoRS variants (introduced as strong baselines in this work) are highlighted in gray.}
	\vspace{-3mm}
	\label{table:flicker_200}
	\resizebox{\textwidth}{!}{ 
		\begin{tabular}{c|c|c|ccc|ccc|ccc|ccc}
		\toprule[1.5pt]
		\multirow{3}{*}{Pairs}&\multirow{3}{*}{Noise}&\multirow{3}{*}{Methods}&\multicolumn{6}{c|}{Flickr30K}&\multicolumn{6}{c}{MS-COCO 1K}\\
		&&&\multicolumn{3}{c|}{Image$\longrightarrow$Text}&\multicolumn{3}{c|}{Text$\longrightarrow$Image}&\multicolumn{3}{c|}{Image$\longrightarrow$Text}&\multicolumn{3}{c}{Text$\longrightarrow$Image}\\
		\cline{4-15}
			&&&R@1&R@5&R@10&R@1&R@5&R@10&R@1&R@5&R@10&R@1&R@5&R@10\\
			\midrule
			\multirow{10}{*}{200}&\multirow{2}{*}{0\%}&LoRS& 14.5   & 38.7  & 53.4 & 8.6 & 25.3  & 36.6   & 4.3 & 14.2  & 22.6 & 2.4 & 9.3 & 15.5 \\
			&&Ours& \textbf{18.4}   & \textbf{44.0} & \textbf{56.6} & \textbf{10.2} & \textbf{30.3}  & \textbf{42.3}   & \textbf{6.0} & \textbf{20.1}  & \textbf{30.4} & \textbf{3.6} & \textbf{12.6} & \textbf{20.5} \\
			\cmidrule{2-15}
			&\multirow{4}{*}{30\%}&LoRS& 12.5   & 32.6  & 48.4 & 7.0 & 22.3  & 33.1  & 3.2 & 11.4  & 18.9 & 1.8 & 6.7 & 11.8 \\
			&&LoRS-C&\cellcolor{tabhighlight}12.5   &\cellcolor{tabhighlight}36.6  &\cellcolor{tabhighlight}51.1 &\cellcolor{tabhighlight}8.0 &\cellcolor{tabhighlight}24.1  &\cellcolor{tabhighlight}35.9   &\cellcolor{tabhighlight}3.7 &\cellcolor{tabhighlight}13.5  &\cellcolor{tabhighlight}22.0 &\cellcolor{tabhighlight}2.2 &\cellcolor{tabhighlight}8.1 &\cellcolor{tabhighlight}14.0 \\
			&&LoRS+&\cellcolor{tabhighlight}12.3   &\cellcolor{tabhighlight}35.9  &\cellcolor{tabhighlight}49.3 &\cellcolor{tabhighlight}7.8 &\cellcolor{tabhighlight}23.6  &\cellcolor{tabhighlight}35.0   &\cellcolor{tabhighlight}3.4 &\cellcolor{tabhighlight}13.2  &\cellcolor{tabhighlight}21.5 &\cellcolor{tabhighlight}1.9 &\cellcolor{tabhighlight}7.9 &\cellcolor{tabhighlight}13.7\\
			&&Ours&\cellcolor{tabhighlight}\textbf{17.4}   &\cellcolor{tabhighlight}\textbf{41.6}  &\cellcolor{tabhighlight}\textbf{54.6} &\cellcolor{tabhighlight}\textbf{9.3} &\cellcolor{tabhighlight}\textbf{27.8}  &\cellcolor{tabhighlight}\textbf{40.2}  &\cellcolor{tabhighlight}\textbf{5.5} &\cellcolor{tabhighlight}\textbf{18.2}  &\cellcolor{tabhighlight}\textbf{28.4} &\cellcolor{tabhighlight}\textbf{3.2} &\cellcolor{tabhighlight}\textbf{11.7} &\cellcolor{tabhighlight}\textbf{19.1}\\
			\cmidrule{2-15}
			&\multirow{4}{*}{50\%}&LoRS& 10.5   & 28.6  & 43.7 & 6.5 & 20.4  & 31.7  & 2.4 & 10.3  & 16.9 & 1.1 & 5.1 & 9.3 \\
			&&LoRS-C&\cellcolor{tabhighlight}12.1   &\cellcolor{tabhighlight}34.6  &\cellcolor{tabhighlight}50.7 &\cellcolor{tabhighlight}7.7 &\cellcolor{tabhighlight}23.3  &\cellcolor{tabhighlight}35.1  &\cellcolor{tabhighlight}3.3 &\cellcolor{tabhighlight}12.2  &\cellcolor{tabhighlight}20.4 &\cellcolor{tabhighlight}1.9 &\cellcolor{tabhighlight}7.4 &\cellcolor{tabhighlight}12.8\\
			&&LoRS+&\cellcolor{tabhighlight}11.0   &\cellcolor{tabhighlight}34.8  &\cellcolor{tabhighlight}49.8 &\cellcolor{tabhighlight}7.0 &\cellcolor{tabhighlight}21.9  &\cellcolor{tabhighlight}32.8  &\cellcolor{tabhighlight}3.4 &\cellcolor{tabhighlight}12.1  &\cellcolor{tabhighlight}20.2 &\cellcolor{tabhighlight}1.7 &\cellcolor{tabhighlight}6.6 &\cellcolor{tabhighlight}11.3 \\
            &&Ours&\cellcolor{tabhighlight}\textbf{16.6}   &\cellcolor{tabhighlight}\textbf{39.5}  &\cellcolor{tabhighlight}\textbf{53.3} &\cellcolor{tabhighlight}\textbf{8.8} &\cellcolor{tabhighlight}\textbf{26.4}  &\cellcolor{tabhighlight}\textbf{39.0} &\cellcolor{tabhighlight}\textbf{5.1} &\cellcolor{tabhighlight}\textbf{16.8}  &\cellcolor{tabhighlight}\textbf{26.4} &\cellcolor{tabhighlight}\textbf{3.0} &\cellcolor{tabhighlight}\textbf{11.3} &\cellcolor{tabhighlight}\textbf{18.6} \\
			\bottomrule[1.5pt]

			\multirow{10}{*}{500}&\multirow{2}{*}{0\%}&LoRS& 15.5   & 39.8  & 53.7 & 10.0 & 28.9  & 41.6 & 5.3 & 18.3  & 27.9 & 2.8 & 9.9 & 16.5 \\
			&&Ours& \textbf{18.9}   & \textbf{48.1} & \textbf{62.2} & \textbf{11.7} & \textbf{32.3}  & \textbf{45.5}  & \textbf{7.5} & \textbf{22.7}  & \textbf{32.8} & \textbf{4.3} & \textbf{14.9} & \textbf{23.4} \\
			\cmidrule{2-15}
			&\multirow{4}{*}{30\%}&LoRS& 12.7   & 35.9  & 49.4 & 7.6 & 23.2  & 35.1 & 3.5 & 13.3  & 22.0 & 2.3 & 8.8 & 14.5 \\
			&&LoRS-C&\cellcolor{tabhighlight}14.3   & \cellcolor{tabhighlight}38.4  & \cellcolor{tabhighlight}52.2 & \cellcolor{tabhighlight}8.8 & \cellcolor{tabhighlight}26.0  & \cellcolor{tabhighlight}38.5 & \cellcolor{tabhighlight}4.2 & \cellcolor{tabhighlight}16.3  & \cellcolor{tabhighlight}25.3 & \cellcolor{tabhighlight}2.6 & \cellcolor{tabhighlight}9.1 & \cellcolor{tabhighlight}15.3\\
			&&LoRS+& \cellcolor{tabhighlight}13.7   & \cellcolor{tabhighlight}37.5  & \cellcolor{tabhighlight}51.9 & \cellcolor{tabhighlight}8.7 & \cellcolor{tabhighlight}25.6  & \cellcolor{tabhighlight}37.4 & \cellcolor{tabhighlight}4.1 & \cellcolor{tabhighlight}15.3  & \cellcolor{tabhighlight}24.9 & \cellcolor{tabhighlight}2.4 & \cellcolor{tabhighlight}9.1 & \cellcolor{tabhighlight}15.6 \\
			&&Ours& \cellcolor{tabhighlight}\textbf{18.0}   &\cellcolor{tabhighlight}\textbf{46.0}  &\cellcolor{tabhighlight}\textbf{59.7} & \cellcolor{tabhighlight}\textbf{10.6} & \cellcolor{tabhighlight}\textbf{29.6}  & \cellcolor{tabhighlight}\textbf{42.6} & \cellcolor{tabhighlight}\textbf{6.7} & \cellcolor{tabhighlight}\textbf{21.6}  & \cellcolor{tabhighlight}\textbf{31.9} & \cellcolor{tabhighlight}\textbf{3.8} & \cellcolor{tabhighlight}\textbf{12.9} & \cellcolor{tabhighlight}\textbf{20.5}\\
			\cmidrule{2-15}
			&\multirow{4}{*}{50\%}&LoRS& 11.2   & 33.5  & 46.0 & 6.0 & 20.5  & 31.0  & 3.7 & 12.7  & 20.3 & 1.5 & 6.0 & 10.5 \\
			&&LoRS-C&\cellcolor{tabhighlight}13.6   & \cellcolor{tabhighlight}35.9  & \cellcolor{tabhighlight}50.8 & \cellcolor{tabhighlight}8.1 & \cellcolor{tabhighlight}25.4  & \cellcolor{tabhighlight}37.8  & \cellcolor{tabhighlight}4.0 & \cellcolor{tabhighlight}14.9  & \cellcolor{tabhighlight}24.1 & \cellcolor{tabhighlight}2.2 & \cellcolor{tabhighlight}8.4 & \cellcolor{tabhighlight}14.2 \\
			&&LoRS+&\cellcolor{tabhighlight}12.6   & \cellcolor{tabhighlight}36.5  & \cellcolor{tabhighlight}50.3 & \cellcolor{tabhighlight}7.8 & \cellcolor{tabhighlight}24.1  & \cellcolor{tabhighlight}35.2 & \cellcolor{tabhighlight}3.9 & \cellcolor{tabhighlight}14.8  & \cellcolor{tabhighlight}23.7 & \cellcolor{tabhighlight}2.0 & \cellcolor{tabhighlight}7.8 & \cellcolor{tabhighlight}13.5 \\
            &&Ours& \cellcolor{tabhighlight}\textbf{16.9}   & \cellcolor{tabhighlight}\textbf{43.4}  &\cellcolor{tabhighlight}\textbf{57.9} & \cellcolor{tabhighlight}\textbf{10.0} & \cellcolor{tabhighlight}\textbf{28.9}  & \cellcolor{tabhighlight}\textbf{40.9}  & \cellcolor{tabhighlight}\textbf{6.2} & \cellcolor{tabhighlight}\textbf{20.2}  & \cellcolor{tabhighlight}\textbf{30.7} & \cellcolor{tabhighlight}\textbf{3.3} & \cellcolor{tabhighlight}\textbf{12.2} & \cellcolor{tabhighlight}\textbf{19.7} \\
			\bottomrule[1.5pt]
	\end{tabular}}
 \vspace{-6mm}
\end{table*}

  \begin{wraptable}{r}{0.5\textwidth}
    \centering
    \vspace{-5mm}
    \caption{Distilling 30\% noise Flickr30K into 1K/2K samples.
    Model on original clean dataset training achieves R@1/5/10 of 33.9/65.1/75.2 (Image) and 27.3/57.1/69.7 (Text) on Flickr30K.}
    \vspace{-2mm}
    \label{table:flicker_1000}
    \resizebox{0.5\textwidth}{!}{
      \begin{tabular}{c|c|ccc|ccc}
      \toprule[1.5pt]
      \multirow{3}{*}{Pairs}&\multirow{3}{*}{Methods}&\multicolumn{6}{c}{Flickr30K}\\
      &&\multicolumn{3}{c|}{Image$\longrightarrow$Text}&\multicolumn{3}{c}{Text$\longrightarrow$Image}\\
      \cline{3-8}
        &&R@1&R@5&R@10&R@1&R@5&R@10\\
        \midrule
        \multirow{2}{*}{1000}&LoRS& 13.0   & 36.5  & 50.2 & 8.1 & 24.6  & 35.7  \\
        &Ours& \textbf{19.2}   & \textbf{49.1} & \textbf{63.0} & \textbf{12.5} & \textbf{32.2}  & \textbf{45.8}  \\
        \midrule
        \multirow{2}{*}{2000}&LoRS& 12.5   & 35.5  & 49.7 & 8.0 & 23.8  & 35.2  \\
        &Ours& \textbf{20.5}   & \textbf{51.2} & \textbf{64.4} & \textbf{14.0} & \textbf{38.4}  & \textbf{51.3}  \\
        \bottomrule[1.5pt]
    \end{tabular}}
   \vspace{-6mm}
  \end{wraptable}

  \section{More Scalability and Practicality Results on COCO and Flickr30K}
  \label{sec: sample_size}
  
  \Cref{table:flicker_200,table:flicker_1000} present the distillation results on noisy Flickr30k and COCO datasets with various noise ratios and distilled dataset sizes.
  In general, the larger distilled datasets capture more knowledge from the real data and consistently improve the performance of all methods.
  Notably, the performance gap between our MDW and the strong baselines is more pronounced as the noise ratio and distilled dataset size increases.
  This is because, as the distilled dataset size grows, the target multi-modal correspondence patterns shift from simple to more complex \cite{guotowards}.
  However, due to its noise sensitivity, LoRS tends to overfit to PMPs during the late training stage when learning harder patterns \cite{arpit2017closer}.
  Conversely, our MDW shows appealing scalability, attributed to its fine-grained correspondence-enhanced distillation and certified noise robustness. 
  Moreover, our MDW demonstrates remarkable scalability, achieving ~74\% full clean data performance with only 2K distilled samples (less than 1\% of the original dataset),  highlighting its appealing practicality for real-world applications.

\section{Efficiency Analyses of MDW}
\label{sec: efficiency}

Efficiency is crucial for large-scale multi-modal dataset distillation. 
We analyze the efficiency of our MDW from two perspectives: robust multi-modal correspondence prior extraction and fine-grained correspondence-enhanced distillation. Overall, our MDW incurs negligible memory and computational overhead while yielding substantial performance improvements, as detailed below.

\begin{wraptable}{r}{0.55\textwidth}
  \centering
  \vspace{-5mm}
  \caption{Efficiency analyses on CC104K, where ``data loss'' denotes training samples discarded due to noise.}
  \vspace{-3mm}
  \label{tab: efficiency}
  \setlength{\tabcolsep}{2.5pt}
  \begin{tabular}{c|cccc}
  \toprule
  Methods & Data Loss & Filtration & Training & Memory \\
  \midrule
  LoRS & - & 0s & 348s & $\sim$12.5 GB\\
  LoRS+ & high & 357s & 255s  &  $\sim$13.0 GB\\
  Ours & low& 21s & 373s & $\sim$14.5 GB\\
  \bottomrule
  \end{tabular}
  \vspace{-6mm}
  \end{wraptable}

\textit{\textbf{Robust Multi-Modal Prior Knowledge Extraction}}: Compared to strong baselines \cite{lors,vl_distill}, MDW leverages expert model to extract robust multi-modal prior knowledge from \textit{\textbf{noisy}} real-world data, with the primary computational cost arising from handling PMPs.
\Cref{tab: efficiency} presents the training overhead on CC104K using an NVIDIA Tesla L40 48G, where our MDW's efficiency mainly falls into two parts: (1) \textbf{\textit{Sample filtration}}: Compared to the baseline LoRS, LoRS+ requires a two-stage process, where a PMP-robust model is first trained for sample filtration before the actual training begins.
Conversely, MDW introduces negligible filtration cost: local-level sample filtration applies a maximum similarity constraint derived as a by-product of the similarity matrix during training, while global-level sample filtration fits a BMM to per-sample similarity distribution, a lightweight step that only takes $\sim$20 seconds.
(2) \textbf{\textit{Sample utilization}}:
To ensure clean subset purity, LoRS+'s data filtration strategy leads to severe information loss, \eg, if 30\% of the $N$ image-text pairs are identified as noisy, the resulting filtered clean dataset contains only $0.7N\times0.7N=0.49N^2$ pairs, retaining merely 49\% of the original.
In contrast, beyond conventional correspondence learning, MDW complementarily exploits non-correspondence information from negative matches across all samples, effectively increasing the number of usable samples to $N^2 - 0.3N$.
Notably, our MDW alleviates real data's information loss without extra cost, as it focus on negative matches with the matching probability matrix (by-products of conventional image-text contrastive learning).

\begin{wraptable}{r}{0.5\textwidth}
  \centering
  \vspace{-4mm}
  \caption{Computational and memory analyses on real-world noise CC104K with 100 distilled samples, with results averaged over 100 iterations.}
  \vspace{-3mm}
  \label{tab: flop}
  \setlength{\tabcolsep}{2.5pt}
  \begin{tabular}{c|cc|cc}
  \toprule
  \multirow{2}{*}{Methods} & \multicolumn{2}{c|}{CEO} & \multicolumn{2}{c}{Traj Matching} \\
  & FLOPs & Memory & FLOPs & Memory \\
  \midrule
  LoRS & - &  - & $\sim$2100G & $\sim$23 GB \\
  Ours & 5G & $\sim$19 MB & $\sim$2100G & $\sim$23 GB \\
  \bottomrule
  \end{tabular}
  \vspace{-3mm}
  \end{wraptable}

\textit{\textbf{Fine-Grained Correspondence-Enhanced Distillation}}: MDW benefits from existing efficient distillation strategies, \eg, low-rank matrix decomposition \cite{lors} and TESLA \cite{cui2023scaling}. 
Compared to the strong baseline LoRS, \Cref{tab: flop} shows that our fine-grained correspondence-enhanced distillation maintains the efficiency of expert-student trajectory matching, as it only applies adaptive weighting to pixel-level gradients of distilled image.
The primary overhead lies in correspondence-enhanced optimization (CEO), including correspondence-discriminative regions identification and associated adaptive weights storage. 
Specifically, we employ Grad-CAM \cite{selvaraju2017grad} to identify such regions, incurring an inference overhead of $\sim$5 GFLOPs per image.
Moreover, the memory cost for storing adaptive weights is negligible, \ie, $O(NHW)$, where $N$ denotes distilled data's size and $H, W$ are distilled images' height and width.

\section{Analyses of Dual-Track Collaboration between $L_c$ and $L_n$}
\label{sec: dual_track}
In our MDW, the collaboration between the correspondence learning track ($L_c$) and the non-correspondence learning track ($L_n$) manifests in both explicit and implicit ways.
(1) Explicitly, $L_c$ and $L_n$ are naturally \textit{\textbf{complementary}}: $L_c$ focuses on correspondence learning using \textit{positive matches} after sample filtration, while $L_n$ targets non-correspondence learning with \textit{negative matches} across all samples.
(2) Implicitly, the noise tolerance of $L_n$ enables more reliable exploitation of negative matches, which in turn strengthens the expert model. This facilitates more accurate filtration of noisy samples (see \Cref{sec: PMP_filtration}), thereby collaboratively enhancing the correspondence learning.

\begin{wraptable}{r}{0.5\textwidth}
	\centering
  \vspace{-4mm}
	\caption{Ablation of robust multi-modal prior knowledge extraction on 30\% noise Flickr30K. Acc. denotes clean sample selection accuracy.}
  \vspace{-2mm}
	\label{tab: sample_selection}
	\resizebox{0.5\textwidth}{!}{ 
		\begin{tabular}{c|c|c|c}
		\toprule[1.5pt]
		\multirow{2}{*}{Methods}&Image$\longrightarrow$Text&Text$\longrightarrow$Image&\multirow{2}{*}{Acc.}\\
		\cline{2-3}
		&	 R@1/5/10 & R@1/5/10 & \\
			\midrule
    	$L_c$ & 29.5/59.1/73.2 & 22.9/50.7/62.6 & 98.7\\
    	$L_n$ & 29.1/60.4/73.6 & 22.2/50.8/63.8 & 99.1\\ 
    	$L_c+L_n$ & 32.0/64.4/78.1 & 25.9/55.2/69.6 &99.7\\
			\bottomrule[1.5pt]
     \end{tabular} 
  }
\end{wraptable}

To validate this analysis, we conduct ablation studies on the Flickr30K dataset with 30\% noise, as shown in \Cref{tab: sample_selection}. Specifically, during robust multimodal prior knowledge extraction from noisy real data, we adopt three expert training objectives: (1) $L_c$, (2) $L_n$ and (3) $L_c + L_n$. As the quality of multimodal prior knowledge extraction is difficult to quantify directly, we use test set accuracy as a proxy and additionally report the corresponding sample selection accuracy.
Using only $L_c$, the filtration process leads to severe information loss, resulting in a notable drop in expert efficacy, with R\_sum drop by $\sim$10\%. When using only $L_n$, although only negative samples utilization support prior knowledge extraction with certifiable noise tolerance, the neglect of informative positive matches yields sub-optimal results, \eg, low R@1 scores across both modalities. 
In contrast, combining $L_c$ and $L_n$ substantially improves expert efficacy and increases sample selection accuracy to 99.7\%, highlighting the necessity of our MDW's dual-track collaboration for robust multimodal prior knowledge extraction.

\section{More Visualization on Distilled Data}
\label{sec: more_distilled_sample_vis}

\begin{figure}[p]
  \centering
  \includegraphics[width=\linewidth]{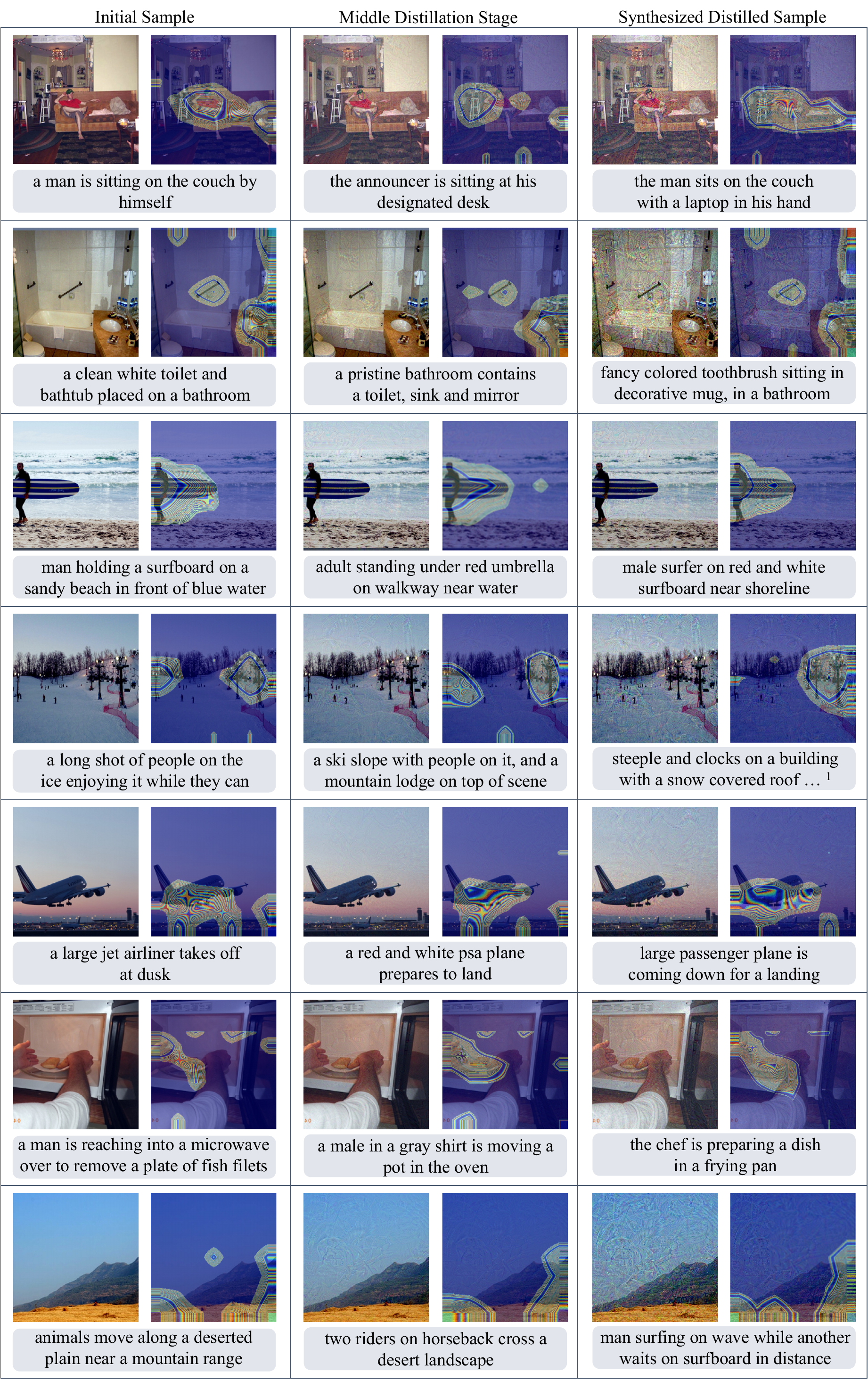}
  \vspace{-5mm}
\caption{More synthetic samples with correspondence-discriminative regions. The full caption of $^1$ is ``steeple and clocks on a building with a snow covered roof with a neon sign in the distance''.}
\vspace{-5mm}
\label{fig: distilled_sample_appendix}
\end{figure}

\Cref{fig: distilled_sample_appendix} shows more initial and distilled image-text pairs on 30\% noise COCO, which further validates the efficacy of our fine-grained correspondence-enhanced distillation in capturing and refining multi-modal correspondence patterns within the distilled images. 
Notably, in the first-row example, the distilled image preserves the overall semantic content while introducing finer details, such as the emergence of a laptop pattern in the person's hands. 
In the last-row example, the distilled image even generates patterns entirely absent in the original input, such as transforming a desert and sky scene into a seaside landscape with additional surfing man patterns. 
These evolved details significantly enrich the information density and utility of the distilled data, thereby enhancing the generalization capability of models trained on it.

\section{More Implementation Details.}
\label{sec: hyperparameter}
\begin{wraptable}{r}{0.5\textwidth}
  \centering
  \vspace{-18mm}
  \caption{Hyperparameters configuration for different datasets and experimental settings, where LR: denote the learning rate and $\tilde{\mathcal{P}}$ denotes the learnable soft matching probabilities.}
  \resizebox{0.5\textwidth}{!}{
  \begin{tabular}{c|ccc|ccc}
  \hline
  \multirow{2}{*}{\textbf{Parameters}} &  \multicolumn{3}{c|}{\textbf{Flickr30K/CC104K}} & \multicolumn{3}{c}{\textbf{COCO}} \\
  \cline{2-7}
   & \textbf{100} & \textbf{200} & \textbf{500} & \textbf{100} & \textbf{200} & \textbf{500} \\
  \hline
  LR: Image                        & 100  & 1000  & 1000  & 1000  & 1000  & 5000 \\
  LR: Text                         & 100  & 1000  & 1000  & 1000  & 1000  & 5000 \\
  LR: LR                           & 0.01 & 0.01  & 0.01  & 0.01  & 0.01  & 0.01 \\
  LR: $\tilde{\mathcal{P}}$                   & 10   & 10    & 100   & 5     & 50    & 500  \\
  Initial LR                       & 0.1  & 0.1   & 0.1   & 0.1   & 0.1   & 0.1  \\
  Batch Size                       & 20   & 20    & 40    & 20    & 20    & 40   \\
  \hline
  Max Start Epoch             & 10    & 10     & 10     & 10     & 10     & 10    \\
  Expert Steps $T_1$                 & 1    & 1     & 1     & 1     & 1     & 1    \\
  Synth Steps $T_2$                     & 8    & 10     & 12     & 8     & 10     & 12    \\
  \hline
  $\delta$                    & 0.6    & 0.6     & 0.6     & 0.6     & 0.6     & 0.6    \\

  \hline
\end{tabular}}
\vspace{-35mm}
  \label{tab: hyperparameter}
  \end{wraptable}
  
  We showcase the hyperparameters configuration of our MDW in \Cref{tab: hyperparameter}. Most distillation parameters are kept the same as previous work \cite{lors,vl_distill} for fairness.

\newpage

\section{Algorithm}
\label{sec: algo}
We have provided the algorithm of our robust multi-modal prior knowledge extraction from noisy real data (\Cref{algo: expert},) fine-grained correspondence-enhanced distillation (\Cref{algo: distillation}) and training a network with our MDW synthetic distilled data (\Cref{algo: inference}).
\begin{algorithm}[!th]
    \caption{\small Algorithm of our robust multi-modal prior knowledge extraction from noisy real data.}
    \label{algo: expert}
    \begin{small}
            
    \KwIn{A noisy multi-modal dataset $\mathcal{D}_\eta$, clean probability threshold $\delta$, expert model $\mathcal{M}$}
    \KwOut{Trained expert model $\mathcal{M}$}
    Warm up the model $\mathcal{M}$ using $L_n$ (Eq. \ref{eq: l_n}).

    \For{$i=1:num\_epochs$}
    {
    \textcolor{blue}{//Consensus-Driven Sample Selection: filtering PMPs with neural networks' memorization effect.}

    \textcolor{blue}{//Global-Level: modeling per-sample similarity distribution with Beta Mixture Model (BMM).}
    
    $\mathcal{S} = \{s_i=h(V_i,T_i)|(V_i,T_i)\in \mathcal{D}_\eta\}$ \hfill \textcolor{blue}{//Per-sample similarity.}
    
    $\mathcal{P}_{\text{clean}} = \{p_i \mid p_i = p\left(k \mid s_i\right), \forall s_i \in \mathcal{S}\} \leftarrow BMM(\mathcal{S})$ \hfill \textcolor{blue}{//Sample-wise clean posterior prob.}
    
    $\mathcal{D}_{\text{clean}}^{\text{global}} = \{(V_i,T_i)|p_i>\delta, \forall p_i\in\mathcal{P}_{\text{clean}}\}$ \hfill \textcolor{blue}{//Global-level clean sample.}

    \For{$j=1:num\_steps$}
    {
        Sample a mini-batch $(I_i,T_j)_{i,j=1}^B$ from $\mathcal{D}_\eta$
        
        Compute the similarity matrix $H_{ij} = h(I_i,T_j)$ of the mini-batch.
        
        \textcolor{blue}{//Local-Level: maximal similarity constraint.}
        
        $\mathcal{D}_{\text{clean}}^{\text{local}} = \{(V_i,T_i)|i=\arg\max\nolimits_jH_{ij}\ \wedge \ i=\arg\max\nolimits_jH_{ji}\}$ \hfill \textcolor{blue}{//Local-level clean sample.}
        
        $\mathcal{D}_{\text{clean}}= \mathcal{D}_{\text{clean}}^{\text{global}}\cap\mathcal{D}_{\text{clean}}^{\text{local}}$ \hfill \textcolor{blue}{//Final clean sample driven by consensus.}
        
        \textcolor{blue}{//Dual-Track Collaborative Learning.}

        Multi-modal correspondence learning on $\mathcal{D}_{\text{clean}}$ with $L_c$.

        Multi-modal non-correspondence learning on negative matches across all samples with $L_n$.

        Update expert model $\mathcal{M}$ with $L_c$ and $L_n$.

    }
    }
\end{small}
\end{algorithm}

\begin{algorithm}[!th]
    \caption{\small Algorithm of our fine-grained correspondence-enhanced distillation.}
    \label{algo: distillation}
    \begin{small}
            
    \KwIn{Noisy multi-modal dataset $\mathcal{D}_\eta$, learning rate $\alpha_v$, $\alpha_t$, $\alpha_p$, interval $T_1$, $T_2$ and $k$.}
    \KwOut{Compact clean distilled data $\tilde{\mathcal{D}}=(\tilde{V}_i, \tilde{T}_j)_{i,j=1}^M$ and soft matching probability matrix $\tilde{\mathcal{P}}$.}
    
    Random distilled sample initialization from $\mathcal{D}_{\text{clean}}$ (Eq. \ref{eq: clean_data}), $\tilde{\mathcal{P}}=\mathbf{I}$
    
    \For{$j=1:num\_steps$}
    {
        \textcolor{blue}{//Correspondence-discriminative regions identification and adaptive weight generation.}
        
        \If{$j \bmod k = 0$}{
            $A^{i} \leftarrow \text{Grad-CAM}(\tilde{V}_i, \tilde{T}_i)$
        }

        Initialize expert and student model parameters $\theta$ and $\tilde{\theta}$ with any initial value, $\theta_0 = \tilde{\theta}_0$.

        Train expert model for $T_1$ steps using Algorithm \ref{algo: expert} on $\mathcal{D}_\eta$ \hfill \textcolor{blue}{//Expert trajectory computation.}

        Train student model for $T_2$ steps using $L_c$ and $L_n$ on $\tilde{\mathcal{D}}$ \hfill \textcolor{blue}{//Student trajectory computation.}

        $L={\|\tilde{\theta}_{T_2} - \theta_{T_1} \|^2}/{\|\theta_{0}-\theta_{T_1} \|^2}$ \hfill \textcolor{blue}{//Expert-Student trajectory matching.}
        
        $\tilde{\mathcal{V}}\leftarrow\tilde{\mathcal{V}}-\alpha_{v}A\odot\frac{\partial L}{\partial \tilde{\mathcal{V}}}, \tilde{\mathcal{T}}\leftarrow\tilde{\mathcal{T}}-\alpha_{t}\frac{\partial L}{\partial \tilde{\mathcal{T}}}, \tilde{\mathcal{P}}\leftarrow\tilde{\mathcal{P}}-\alpha_{p}\frac{\partial L}{\partial \tilde{\mathcal{P}}}$ \hfill \textcolor{blue}{//Ditilled data update with adaptive weight.}
    }
\end{small}
\end{algorithm}

\begin{algorithm}[!th]
    \begin{small}
    \caption{\small Algorithm of training a network with our MDW synthetic distilled data.}
    \label{algo: inference}
            
    \KwIn{Distilled data $\tilde{\mathcal{D}}=(\tilde{V}_i, \tilde{T}_j)_{i,j=1}^M$, soft matching probability matrix $\tilde{\mathcal{P}}$ and random initial model $\mathcal{M}$}
    \KwOut{Trained model $\mathcal{M}$}
    
    \For{$j=1:num\_steps$}
    {
        Sample a mini-batch $(\tilde{V}_i, \tilde{T}_j)_{i,j=1}^B$ from $\tilde{\mathcal{D}}$ and their soft matching probability $\tilde{p}\in\mathbb{R}^{B\times B}$ from $\tilde{\mathcal{P}}$.
        
        Model training on mini-batch with $L=L_c+L_n$.

        Update model $\mathcal{M}$ with $\mathcal{M}\leftarrow \mathcal{M}-\alpha \frac{\partial L}{\partial \mathcal{M}}$.
    }
    
\end{small}
\end{algorithm}

\end{document}